%% file: arxiv_main.tex
\documentclass[11pt]{article} 
\usepackage{fullpage}
\usepackage{natbib}
\usepackage{float}
\usepackage[hyphens]{url}
\usepackage[hyphenbreaks]{breakurl}
\usepackage[colorlinks=true, allcolors=blue]{hyperref}

\input{math_commands}

\usepackage{tabularx,colortbl,booktabs}
\usepackage{enumitem,color}

\usepackage{hyperref}
\usepackage{algorithm}
\usepackage{url}
\usepackage{stackengine}
\usepackage{graphicx}
\usepackage[export]{adjustbox}
\usepackage{booktabs,enumitem}
\usepackage{makecell}
\usepackage{array}
\usepackage{newfloat}
\usepackage[dvipsnames]{xcolor}
\usepackage{siunitx}
\usepackage{subcaption}

\title{High-Dimensional Black-Box Optimization via Guidance Sampling of Diffusion Models}

\author{%
  \begin{tabular}{ccc}
    Xiaoyan Hu\thanks{Robotics Institute, Carnegie Mellon University}
    & Rohit Sonker\footnotemark[1]
    & Jalal Butt\thanks{Princeton University}
    \\[1em]
    Jiayu Chen\thanks{Data and Systems Engineering, University of Hong Kong}
    & Egemen Kolemen\footnotemark[2]\textsuperscript{,}%
      \thanks{Princeton Plasma Physics Laboratory}
    & Jeff Schneider\footnotemark[1]
  \end{tabular}%
}

\begin{document}

\maketitle

\begin{abstract}
\input{0-abs}
\end{abstract}

\section{Introduction}
\label{1-intro}
\input{1-intro}

\section{Preliminaries}
\label{3-preliminaries}
\input{3-preliminaries}

\section{Guidance Sampling for High-Dimensional Black-Box Optimization}
\label{4-alg}
\input{4-alg}

\section{Numerical Results}
\label{5-exp}
\input{5-exp}

\section{Discussion}
\label{6-discussion}
\input{6-discussion}

\bibliographystyle{abbrvnat}
\bibliography{iclr2026_conference}

\newpage
\appendix
\input{appendix}

\end{document}

%% file: math_commands.tex
\usepackage{amsmath,amssymb,amsthm,amsfonts,bm,bbm, nicefrac}
\usepackage{algorithm}
\usepackage{algpseudocode}
\usepackage{booktabs}
\usepackage{graphicx}
\usepackage{subcaption}
\usepackage{siunitx}
\usepackage[dvipsnames]{xcolor}
\usepackage{enumitem}
\usepackage{makecell}
\usepackage{longtable}

\usepackage{multirow}
\usepackage{array}
\usepackage{pifont}

\newcommand{\cmark}{\ding{51}}
\newcommand{\xmark}{\ding{55}}

\definecolor{matplotlibforest}{HTML}{228B22}
\definecolor{matplotpaleviolet}{HTML}{DB7093}
\definecolor{matplotseagreen}{HTML}{2E8B57}
\definecolor{matplotsteelblue}{HTML}{4682B4}
\definecolor{matplotchocolate}{HTML}{D2691E}
\definecolor{matplotmediumpurple}{HTML}{9370DB}
\definecolor{matplotdarkorange}{HTML}{FF8C00}
\definecolor{matplotgoldenrod}{HTML}{DAA520}

\newcommand{\subA}{{{\textcolor{matplotdarkorange}{BReaD-SubA}}}}
\newcommand{\subB}{{{\textcolor{matplotmediumpurple}{BReaD-SubB}}}}
\newcommand{\unguided}{{{\textcolor{matplotlibforest}{unguided}}}}
\newcommand{\mlp}{{{\textcolor{matplotpaleviolet}{MLP RM-guided}}}}
\newcommand{\fwd}{{{\textcolor{matplotsteelblue}{fwd. uni-guided}}}}
\newcommand{\gpucb}{{{\textcolor{matplotchocolate}{DM-GP-UCB}}}}

\def\hq{{{\hat{q}}}}
\def\tq{{{\tilde{q}}}}
\def\hf{{{\hat{f}}}}
\def\ystar{{{y^\star}}}
\def\fstar{{{f^\star}}}

\def\qd{{{q_\textup{data}}}}
\def\qGpk{{{q^k_{\textup{GP},y^\star}}}}

\def\qGpkt{{{q^k_{\textup{GP},y^\star,t}}}}
\def\qGpkT{{{q^k_{\textup{GP},y^\star,T}}}}
\def\qGpktr{{{q^k_{\textup{GP},y^\star,T-t}}}}
\def\dd{{{\mathrm{d}}}}
\def\barX{{{\bar{X}}}}
\def\barXr{{{\bar{X}^\leftarrow}}}
\def\barXrk{{{\bar{X}^{\leftarrow, k}}}}

\def\epsUncond{\epsilon_\textup{uncond}}
\def\op{{{\mathrm{op}}}}
\def\cov{{{\mathrm{Cov}}}}
\def\var{{{\mathrm{Var}}}}
\def\tr{{{\mathrm{Tr}}}}

\def\eqref#1{equation~\ref{#1}}

\def\1{\bm{1}}

\def\vw{{\bm{w}}}
\def\vx{{\bm{x}}}

\DeclareMathAlphabet{\mathsfit}{\encodingdefault}{\sfdefault}{m}{sl}
\SetMathAlphabet{\mathsfit}{bold}{\encodingdefault}{\sfdefault}{bx}{n}

\def\gD{{\mathcal{D}}}
\def\gE{{\mathcal{E}}}

\def\gG{{\mathcal{G}}}
\def\gH{{\mathcal{H}}}

\def\gN{{\mathcal{N}}}

\def\gP{{\mathcal{P}}}

\def\gU{{\mathcal{U}}}

\def\gX{{\mathcal{X}}}

\def\sR{{\mathbb{R}}}

\newcommand{\E}{\mathbb{E}}

\newcommand{\KL}{D_{\mathrm{KL}}}
\newcommand{\TV}{D_{\mathrm{TV}}}

\DeclareMathOperator*{\argmax}{arg\,max}

\theoremstyle{plain}
\newtheorem{theorem}{Theorem}
\newtheorem{proposition}{Proposition}
\newtheorem{lemma}{Lemma}

\theoremstyle{definition}

\newtheorem{assumption}{Assumption}
\theoremstyle{remark}
\newtheorem{remark}{Remark}

%% file: 0-abs.tex
Many real-world optimization problems involve optimizing black-box objectives for which only a limited number of evaluations are available. Bayesian optimization (BO) is a powerful framework for such problems, but its effectiveness often deteriorates in high-dimensional spaces. Recent work leverages diffusion models for black-box optimization, often through pre-training a reward-conditioned score network to steer generation towards high-reward samples. Alternatively, gradient-based guidance methods assume access to a differentiable reward function from which guidance gradients are queried. However, these approaches either require large amounts of labeled data for offline training or access to exact reward gradients, limiting their applicability in black-box settings. We propose \textsc{BReaD} (\textbf{B}lack-box \textbf{Re}ward-guided S\textbf{a}mpling of \textbf{D}iffusion Model), a guidance sampling framework for optimizing black-box rewards without assuming access to a labeled dataset or reward gradients. The proposed method uses a Gaussian process (GP) as a differentiable surrogate for the unknown objective and actively queries labels for the generated samples. We provide a theoretical analysis of the total-variation distance between the generated distribution and the target distribution. Using Upper Confidence Bound~(UCB) as guidance signals, our method balances exploration and exploitation during generation and enables the discovery of high-reward regions in the data space. Numerical experiments on scientific discovery and image generation tasks demonstrate that our framework can effectively optimize black-box rewards in low-data scenarios.


%% file: 1-intro.tex
Optimizing expensive black-box functions arises in many real-world applications such as robotics~\citep{6630691}, material design~\citep{Khatamsaz2023}, and drug discovery~\citep{8539993}. In the black-box setting, the objective function is unknown and can only be evaluated through costly queries. Bayesian optimization~\citep[BO]{NIPS2012_05311655, 7352306, frazier2018tutorialbayesianoptimization, 10.1145/2939672.2945367, Berkenkamp2023} has emerged as a powerful framework for optimizing such black-box functions. Despite its success, BO often struggles to scale to high-dimensional spaces, where modeling the objective and searching for promising query points become increasingly challenging~\citep{9555522, 10.1145/3545611}.

\begin{figure}[!t]
    \centering
    \includegraphics[width=\linewidth]{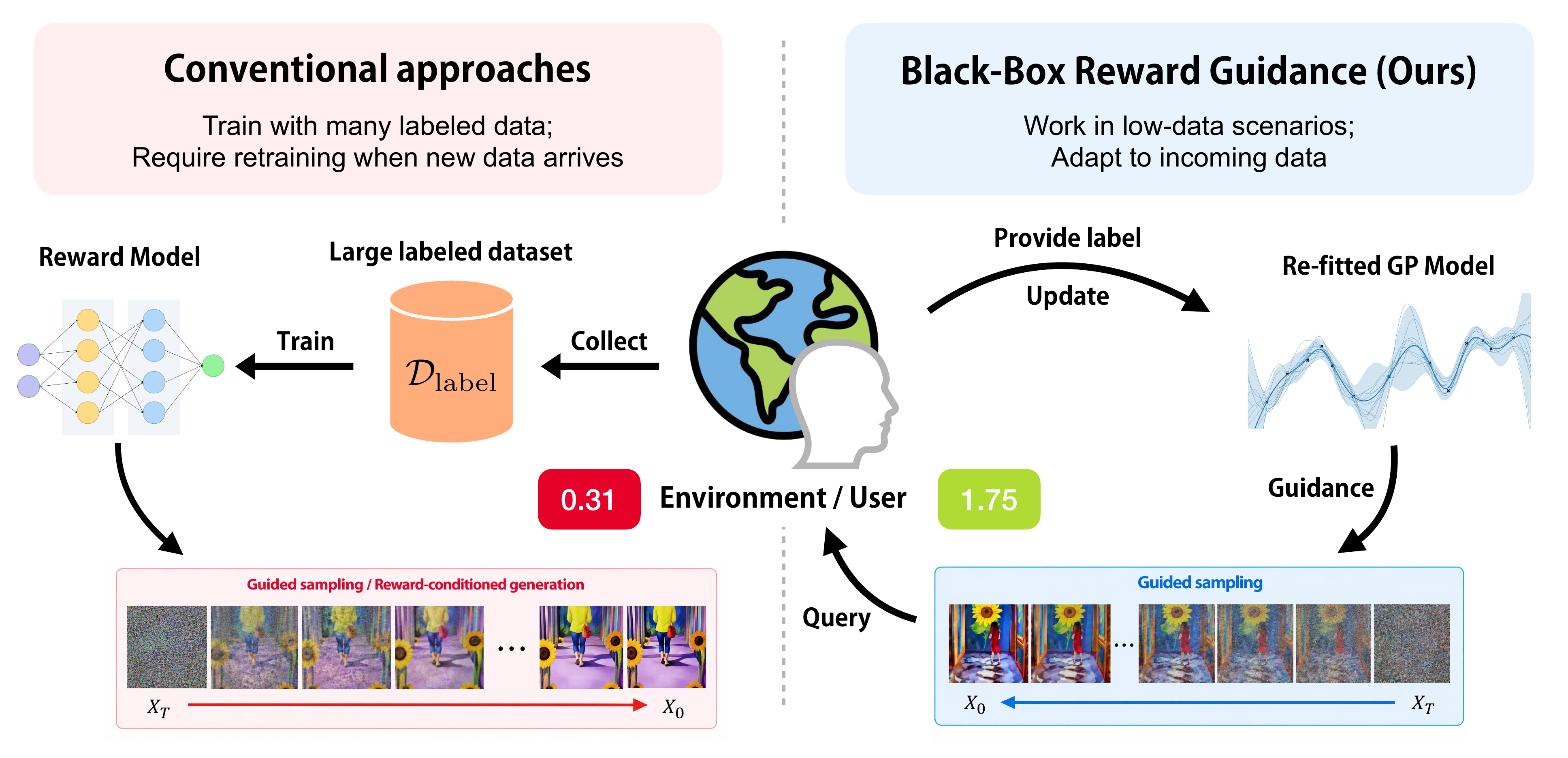}
    \caption{\textbf{Black-box Reward-guided Sampling of Diffusion Models (\textsc{BReaD}).} By actively exploring and querying reward labels in promising regions of the (embedded) sample space, \textsc{BReaD} can effectively optimize black-box rewards in high-dimensional but low-data regimes.}
    \label{fig1}
\end{figure}

Meanwhile, diffusion models~\citep{pmlr-v37-sohl-dickstein15, NEURIPS2020_4c5bcfec} have demonstrated remarkable capabilities in modeling complex data distributions. Controlled generation methods~\citep{dhariwal2021diffusion, ho2021classifierfree, chung2023diffusion, bansal2024universal} further enable sample generation with desired properties, which motivates recent work on diffusion model-based methods for data-driven black-box optimization~\citep{pmlr-v202-krishnamoorthy23a, yuan2023rewarddirected, li2024diffusionmodeldatadrivenblackbox, wu2024diffbbo, pmlr-v258-kong25b, yun2025posterior}. Due to limited space, we defer a detailed discussion of related work to Appendix~\ref{2-related_work}. Existing approaches typically require large amounts of labeled data to learn a reward model or a reward-conditioned score network, making them challenging to apply when only a small number of reward evaluations, e.g., on the order of \num{50}, are available. For example, a user may wish to personalize image generation to their preferences after providing only a few rounds of feedback (Figure~\ref{fig1}). In scientific discovery applications such as nuclear fusion and protein design, labeled data can be scarce and costly to obtain. This motivates the following question: 
\begin{center}
    \textit{Can we optimize black-box rewards in high-dimensional space using only a few evaluations?}
\end{center}

In this paper, we propose \textsc{BReaD}, shorthand for \textbf{B}lack-box \textbf{Re}ward-guided S\textbf{a}mpling of \textbf{D}iffusion Model, a framework for optimizing black-box reward functions in high-dimensional data spaces. \textsc{BReaD} combines a diffusion model that is pre-trained on an \textit{unlabeled} dataset with a Gaussian process (GP) that models the unknown reward function from limited labeled data. We introduce two mechanisms for guided generation from the pre-trained diffusion model. The first, \textit{reward-directed sampling}, uses reward-conditioned gradients derived from the fitted GP to guide generation toward samples that yield a target reward. We provide a theoretical analysis that bounds the total-variation distance~(TV) between the resulting guided distribution and the target distribution. The second, \textit{uncertainty-guided sampling}, incorporates Upper Confidence Bound~(UCB) into guidance to encourage exploration toward sample space with both high predicted rewards or high uncertainty. To effectively model the black-box objectives in high-dimensional data spaces, we leverage deep kernel learning~\citep{wilson2015deepkernellearning} to construct expressive GP kernels over learned low-dimensional representations. We evaluate \textsc{BReaD} on several image generation tasks and simulated fusion plasma control. Our results demonstrate that the proposed framework can effectively optimize black-box rewards with only a small number of labeled samples, achieving performance comparable to guidance methods with access to the exact reward gradient. Moreover, UCB-based exploration improves sample efficiency and enables more effective discovery of high-reward regions.

Our contributions are summarized as follows.
\begin{itemize}[leftmargin=*]
    \item We propose a guidance sampling framework for black-box optimization in high-dimensional spaces. By combining a pre-trained diffusion model with a GP that models the black-box objective, our method effectively generates high-fidelity data with desired properties using only a few rounds of evaluations.
    \item We introduce two subroutines for guiding generation toward high-reward samples. \textit{Reward-directed sampling} uses the GP likelihood conditioned on a target reward as the guidance signal. \textit{Uncertainty-guided sampling} uses the UCB as the guidance signal to balance exploitation and exploration in the sample space.
    \item We provide a theoretical analysis of the total-variation distance between the generated distribution and the target distribution. 
    \item Numerical experiments on image generation tasks and simulated fusion plasma control show that our method can effectively optimize black-box rewards with only a small number of reward evaluations and achieve performance comparable to guidance methods that have access to the exact reward gradients.
\end{itemize}

%% file: 3-preliminaries.tex
\subsection{Score-Based Generative Models}

Given samples from a data distribution $\qd \in \Delta(\gX)$, where $\gX \subset \sR^d$ is the sample space, we aim to learn a model distribution $p_\theta$ that approximates $\qd$ and is easy to sample from. Score-based generative models~\citep[SGMs]{pmlr-v37-sohl-dickstein15, NEURIPS2020_4c5bcfec, song2021scorebased} define a forward SDE from time $0$ to $T$
\begin{equation}
    \dd X_t = - g^2(t) X_t \dd t + \sqrt{2} g(t) \dd B_t, \textup{ where } X_0 \sim \qd.
\label{eq: forward_process}
\end{equation}
Here, $g: \sR_+ \to \sR$ is a positive smooth function, and $(B_t)_{t \ge 0}$ is the standard Wiener process in $\sR^d$. Let $q_t := \textup{law} (X_t)$. We have that $q_t(X_t | X_0) = \gN(X_t; \sqrt{\bar \alpha_t} X_0, (1 - \bar \alpha_t) I)$, where we define $\bar \alpha_t := \exp ( - 2 \int_0^t g^2(s) \dd s)$. Under regularity conditions, the forward process has an equivalent backward SDE~\citep{ANDERSON1982313} from time $0$ to $T$
\begin{equation}
    \dd X_t^\leftarrow = [g^2(T-t) X_t^\leftarrow + 2g^2(T-t) \nabla \log q_{T-t} (X_t^\leftarrow)] \dd t + \sqrt{2} g(T-t) \dd B_t^\leftarrow, \textup{ where }  X_0^\leftarrow \sim q_T,
\end{equation}
where $(B_t^\leftarrow)_{t \in [0, T]}$ is the reversed Brownian motion, and $\nabla \log q_t$ is the \textit{score function} for $q_t$. Particularly, the marginal distribution of $X_t^\leftarrow$ follows $q_{T-t}$ for any $t \in [0, T]$. By Tweedie's formula~\citep{efron2011tweedie}, the posterior mean of $X_0$ conditioned on $X_t^\leftarrow$ is given by
\begin{equation}
    m_{T-t}(X_t^\leftarrow) := \E_{q_0(\cdot|X_t)} [X_0] = \frac{1}{\sqrt{\bar\alpha_{T-t}}} (X_t^\leftarrow + (1 - \bar \alpha_{T-t}) \nabla \log q_{T-t} (X_t^\leftarrow))
\label{eq: posterior_mean}
\end{equation}
To estimate the score function, denoising score matching minimizes the loss
\begin{equation}
    \min_\theta \E_{t \sim \gU([0, T]), X_0 \sim \qd, X_t \sim q_t(\cdot|X_0)} [ w(t) \| s_t^\theta(X_t) - \nabla_{X_t} \log q_t (X_t | X_0) \|^2],
\end{equation}
where $w: \sR_+ \to \sR_+$ is a weighting function. We define the predicted posterior-mean
\begin{equation}
    \hat{x}_0^\theta(x_t^\leftarrow, T-t) = \frac{x_t^\leftarrow + (1 - \bar \alpha_{T-t}) s_{T-t}^\theta(x_t^\leftarrow)}{\sqrt{\bar \alpha_{T-t}}}.
\label{eq: predicted_mean}
\end{equation}

\subsection{Guided Generation from Diffusion Models
}

Given a pre-trained SGM, guided generation aims to generate a sample $X$ with a desired property $Y=y$ from the distribution $P(X|Y=y)$. To this end, we need to learn the conditional score function $\nabla_{x_t} \log q_t (x_t | y)$. By Bayes' rule, we have
\begin{equation}
    \nabla_{x_t} \log q_t (x_t | y) = \nabla_{x_t} \log q_t(x_t) + \nabla_{x_t} \log q_t (y | x_t),
\end{equation}
where the unconditional score can be estimated by the pre-trained score network, i.e., $s_t^\theta (x_t) \approx \nabla_{x_t} \log q_t(x_t)$. To estimate $\nabla_{x_t} \log q_t (y | x_t)$, classifier guidance~\citep{dhariwal2021diffusion} trains auxiliary classifiers on noisy samples $x_t$ with discrete labels and then uses the gradient of of its log conditional probability as the guidance score. Since such classifiers are domain-specific and generally not readily available, classifier-free guidance~\citep[CFG]{ho2021classifierfree} instead jointly trains conditional and unconditional diffusion models and combines their score estimates during generation. However, CFG requires the diffusion model to be trained with the desired conditioning information and does not readily accommodate new guidance modalities without retraining. To address this issue, diffusion posterior sampling~\citet[DPS]{chung2023diffusion} and forward universal guidance~\citep[{Algorithm~\ref{alg:fwd-uni-guidance}}]{bansal2024universal} approximate the conditional score by
\begin{equation}
    \nabla_{x_t} \log q_t (y | x_t) \approx \nabla_{x_t} \log q_0 (y | m_t(x_t)).
\end{equation}

\begin{algorithm}[!t]
\caption{\textsc{ForwardUniversalGuidance}~\citep{bansal2024universal}}
\label{alg:fwd-uni-guidance}
\begin{algorithmic}[1]
\Require guidance function $f: \gX \to \sR$, guidance strength $\eta: \sR_+ \to \sR_+$
\State Sample $X_0^\leftarrow \sim \gN(0, I)$.
\For{denoising step $k = 1, \cdots, \lfloor \frac{T}{h} \rfloor$}
    \State Sample
        \[
            \dd X_t^\leftarrow = [g^2(T-kh) X_{kh}^\leftarrow + 2g^2(T-kh) \tilde s_{T-kh}(X_{kh}^\leftarrow) ] \dd t + \sqrt{2} g(T-{kh}) \dd B_{kh}^\leftarrow,
        \]
    \Statex \hspace{\algorithmicindent}for $t \in [kh, (k+1)h)$, where
        \begin{gather}
            \tilde s_{T-t}(x_t^\leftarrow)
            =
            s_{T-t}^\theta(x_t^\leftarrow)
            +
            \eta(t) \nabla_{x_t^\leftarrow}
            f\left( \hat{x}_0^\theta(x_t^\leftarrow, T - t) \right). 
        \end{gather}
\EndFor
\State \Return $X_T^\leftarrow$
\end{algorithmic}
\end{algorithm}

\subsection{Bayesian Optimization}

Let $\gX \subset \sR^d$ be a compact domain. Bayesian optimization~\citep[BO]{frazier2018tutorialbayesianoptimization} aims to solve $\max_{x \in \gX} f(x)$, where $f: \gX \to \sR$ is the objective function. BO consists of two main components: a Bayesian statistical model for modeling the objective function, and an acquisition function for deciding where to sample next. One common choice for the Bayesian statistical model is the Gaussian Process~\citep[GP]{10.7551/mitpress/3206.001.0001}. A $\gG\gP (\mu(x), \kappa(x, x^\prime))$ is specified by its mean function $\mu(x) = \E[f(x)]$ and covariance (or kernel) function $\kappa(x, x^\prime) = \E[(f(x) - \mu(x)) (f(x^\prime) - \mu(x^\prime)]$. Without loss of generality, we assume that $\mu(x) = 0$ for GPs not conditioned on data. Given noisy evaluations $\{ y_n \}_{n=1}^N$ at a collection of points $\{ x_n \}_{n=1}^N$, where $y_n = f(x_n) + \varepsilon_n$ with i.i.d. Gaussian noise $\varepsilon_n \sim \gN(0, \tau^2)$, the conditional distribution of $f(x)$ is a Gaussian distribution $\gN(\mu_N (x), \sigma_N^2(x))$, where
\begin{gather}
    \mu_N(x) = \kappa(x, x_{1:N}) \left(\kappa(x_{1:N}, x_{1:N}) + \tau^2 I \right)^{-1} y_{1:N}, \label{eq: mu_N} \\
    \sigma_N^2(x) = \kappa(x, x) - \kappa(x, x_{1:N}) \left(\kappa(x_{1:N}, x_{1:N}) + \tau^2 I \right)^{-1} \kappa(x_{1:N}, x). \label{eq: sigma_N}
\end{gather}
To sample the next point $x_{N+1}$ for evaluation, the GP-UCB algorithm~\citep{3104322.3104451} proposes to use the Upper Confidence Bound~(UCB) as an acquisition function to select the candidate
\begin{equation}
    x_{N+1} = \argmax_{x \in \gX} \mu_N(x) +  c_N \sigma_N(x),
\end{equation}
where $c_N$ is a hyperparameter. The UCB acquisition function prefers points whose objective value is uncertain (large $\sigma_N(\cdot)$) and expected to be high~(large $\mu_N(\cdot)$). GP-UCB has demonstrated strong performance in low- to moderate-dimensional optimization problems. However, directly optimizing the UCB acquisition function becomes challenging in high-dimensional data spaces. Instead, our method uses UCB to guide exploration toward promising regions of the data space, while leveraging a pre-trained diffusion model to generate high-fidelity samples.

%% file: 4-alg.tex
Existing diffusion model-based black-box optimization methods often rely on a sufficiently large reward-labeled dataset to train a reward model or a reward-conditioned score network. In many real-world applications, however, one often has access to abundant unlabeled data while reward labels must be actively acquired. 
In this section, we propose a guidance sampling framework for high-dimensional black-box optimization.

\subsection{Problem Setup}

The objective is to optimize a black-box function
\begin{equation}
    \argmax_{x \in \gX} \fstar(x), \textup{ where }  \fstar: \gX \to [0,1].
\end{equation}
We consider a high-dimensional data space $\gX$, such as the space of images. Moreover, we assume access to a large unlabeled dataset sampled from the data distribution $\qd$. Importantly, we do \textit{not} assume access to an offline reward-labeled dataset. Rather, reward labels are actively acquired through an online interaction protocol (see the RHS of Figure~\(\ref{fig1}\)). At each iteration $k$, the algorithm generates a sample $x_k$, queries its reward from the environment, and uses the newly acquired pair $(x_k, y_k)$ to update its reward estimator and subsequent sampling strategy. This setting differs from conventional Bayesian optimization~(BO) problems~\citep{frazier2018tutorialbayesianoptimization}, which typically operate in low- to moderate-dimensional search spaces. Directly applying BO methods can be challenging, as modeling the black-box objective and optimizing the acquisition function become increasingly difficult as the dimensionality grows. Our setting also differs from existing diffusion model-based black-box optimization frameworks~\citep{pmlr-v202-krishnamoorthy23a, yuan2023rewarddirected, li2024diffusionmodeldatadrivenblackbox, wu2024diffbbo, pmlr-v258-kong25b}, which assume access to a pre-collected labeled dataset for offline training. 

\subsection{Guided Sampling of Diffusion Models via Black-Box Rewards}

In this section, we introduce the framework of \textbf{B}lack-box \textbf{Re}ward-guided S\textbf{a}mpling of \textbf{D}iffusion Model~(\textsc{BReaD}, Algorithm~\ref{alg:gp-diffusion}). Our approach combines a diffusion model that is pre-trained on an unlabeled dataset with a Gaussian process (GP) that models the black-box objective. The GP helps discover high-reward regions of the data space, and the pre-trained diffusion model serves as prior distribution and generates valid and high-fidelity sample. At each iteration $k$, \textsc{BReaD} refits the GP posterior using the reward-labeled samples observed so far~(Line 2), generates a new sample $x_k$ via guided diffusion sampling~(Lines 3-8), queries its reward $y_k \sim \gN(\fstar(x_k), \tau^2)$ from the environment~(Line 9), and augments the reward-labeled dataset with the newly observed sample-reward pair~(Line 10).

\begin{algorithm}[!t]
\caption{\textbf{B}lack-box \textbf{Re}ward-guided S\textbf{a}mpling of \textbf{D}iffusion Model~(\textsc{BReaD})}
\label{alg:gp-diffusion}
\begin{algorithmic}[1]
\Require Unlabeled dataset $\gD$, reward-labeled dataset $\gD^{\mathrm{reward}}=\emptyset$, black-box objective $\fstar:\gX\to[0,1]$, Gaussian process prior $\gG\gP(0,\kappa)$, target reward $y^\star \in [0, 1]$, UCB parameter $\lambda \in \sR$, guidance strength $\eta: \sR_+ \to \sR_+$

\For{iteration $k=1,2,\cdots$}
    \State Fit the GP posterior $\fstar \mid \gD^{\mathrm{reward}} \sim \gG\gP(\mu_{k}(\cdot),\kappa_{k}(\cdot, \cdot))$.
    \If{Reward-directed Guidance (\textsc{Subroutine A})}
        \State Set $f_k(x) = \log p_\textup{GP}^k(\ystar|x)$.
    \ElsIf{Uncertainty-aware sampling (\textsc{Subroutine B})}
        \State Set $f_k(x) = \mu_k(x) + \lambda \sigma_k(x)$, where $\mu_k$ and $\sigma_k$ are given by Equations~(\ref{eq: mu_N}-\ref{eq: sigma_N}).
    \EndIf
    \State Sample $x_k = \textsc{ForwardUniversalGuidance}(f_k, \eta)$.
    \State Observe reward $y_k \sim \gN(\fstar(x_{k}), \tau^2)$ for the generated sample $x_{k}$.
    \State Update the reward-labeled dataset $\gD^{\mathrm{reward}}
        =
        \gD^{\mathrm{reward}}
        \cup
        \{(x_{k},y_{k})\}$.
\EndFor
\end{algorithmic}
\label{alg}
\end{algorithm}

We consider two complementary strategies for steering generation toward high-reward samples. The first is motivated by reward-conditioned generation~\citep{pmlr-v202-krishnamoorthy23a}, which generates samples conditioned on a target reward value. During optimization, the conditioning value is iteratively updated to the best reward observed so far. The second is motivated by gradient-based guidance~\citep{dhariwal2021diffusion, bansal2024universal}, which uses gradients of a differentiable function as the guidance signal. Since the true gradients are unavailable in our black-box setting, we use the fitted GP as a differentiable surrogate to construct the guidance functions in both strategies. Accordingly, we introduce two subroutines that use forward universal guidance~\citep[Algorithm~\ref{alg}]{bansal2024universal} with different guidance functions derived from the GP.

\textbf{\textsc{Sub-Routine A.}  Reward-conditioned guidance via GP-likelihood.}\quad
This sub-routine aims to generate samples whose rewards are close to a pre-specified target value $\ystar$. Ideally, we would sample from the reward-conditioned target distribution
\begin{equation}
    q_\ystar(x) \propto \qd(x) \gN(\ystar; \fstar(x), \tau^2),
\label{eq: q_ystar}
\end{equation}
where $\qd$ is the data distribution and $\tau^2$ is the observation-noise variance. Given the GP fitted at iteration $k$, we approximate the target distribution by $q_{\textup{GP}, \ystar}^k (x) \propto \qd(x) p_{\textup{GP}}^k (\ystar | x)$, where we denote by $p_{\textup{GP}}^k(\ystar | x)$ the GP predictive likelihood of observing the target reward $\ystar$ for sample $x$. Specifically, let $\mu_k$ and $\sigma_k^2$ be the mean~(\ref{eq: mu_N}) and the variance~(\ref{eq: sigma_N}) of the conditional distribution of $\fstar$, we have $p_{\textup{GP}}^k(\ystar | x) := \gN(\ystar; \mu_k(x), \sigma_k^2(x) + \tau^2)$. To approximately sample from this distribution, we use the guided score
\begin{equation}
    \tilde{s}_{\textup{GP}, T-t}^k(x_t^\leftarrow) = s_{T-t}^\theta (x_t^\leftarrow) + \eta (t) \nabla_{x_t^\leftarrow}
            \log p_{\textup{GP}}^k \left( \ystar | \hat{x}_0^\theta(x_t^\leftarrow, T-t) \right),
\end{equation}
at each time $t$ of the reverse process, where $s_{T-t}^\theta$ is the unconditional score estimated by the pre-trained diffusion model, $\hat{x}_0^\theta(x_t^\leftarrow, T-t)$ is the predicted posterior mean~(\ref{eq: predicted_mean}). We analyze the total-variation~(TV) distance between the generated distribution and the target reward-conditioned distribution. Our analysis decomposes the distribution mismatch into five sources, including GP approximation, score estimation, discretization, posterior averaging, and initialization errors. The GP approximation error arises from the discrepancy between the GP surrogate and the true black-box objective, which decreases as the GP surrogate becomes more accurate with additional reward observations. The remaining errors arise from approximating the reward-conditioned sampling process using forward universal guidance.  The complete bound is provided in Appendix~\ref{sec: B.3}.

\begin{theorem}[Distribution mismatch, simplified]
\label{thm: main}
Let $\hat q_\ystar^k$ denote the sample distribution generated by \textsc{Subroutine A} of Algorithm~\ref{alg:gp-diffusion} at iteration $k$. Under Assumptions~\ref{as: bounded_reward}-\ref{as: der_log_gp} in Appendix~\ref{sec: B.2}, the total-variation distance between $\hat q_{\ystar}^k$ and the target distribution~(\ref{eq: q_ystar}) satisfies
\begin{equation}
\begin{aligned}
    \TV(q_\ystar, \hq^k_{\ystar}) \lesssim \varepsilon_{\textup{GP}, k} + \varepsilon_{\textup{init}} + \varepsilon_{\textup{score}} + \varepsilon_{\textup{disc}} + \varepsilon_{\textup{post-avg}}.
\end{aligned}
\end{equation}
\begin{proof}
    See Appendix~\ref{sec: B}.
\end{proof}
\end{theorem}

\textbf{\textsc{Sub-Routine B.} Uncertainty-aware guidance via UCB.}\quad
To encourage exploration in the data space, this subroutine incorporates the Upper Confidence Bound~\citep[UCB]{10.1023/A:1013689704352} into guidance. We use the guided score
\begin{gather}
    \tilde{s}_{\textup{UCB}, T-t}^k(x_t^\leftarrow) = s_{T-t}^\theta (x_t^\leftarrow) + \eta (t) \nabla_{x_t^\leftarrow} \textup{UCB}_k(x_t^\leftarrow)
\label{eq: ucb-guided}
\end{gather}
at each time $t$ of the reverse process, where $\textup{UCB}_k(x_t^\leftarrow) := \left[ \mu_k + \lambda \sigma_k  \right](\hat{x}_0^\theta(x_t^\leftarrow, T-t))$ is the (predicted) UCB value for the noisy sample, and $\lambda$ controls the trade-off between exploitation and exploration. The guided process~(\ref{eq: ucb-guided}) aims to approximate the data distribution reweighted by the UCB value, i.e., $q_\textup{UCB}^k(x) \propto \qd(x) \exp (\textup{UCB}_k (x))$. In the GP bandit literature~\citep{3104322.3104451}, UCB provides an optimistic estimate of the black-box objective $\fstar$. By steering generation toward the high-UCB region of the data space and querying reward labels for these samples, the GP acquires informative observations that improve its estimation. This further enables the discovery of the promising high-reward region. In numerical experiments, we find that UCB guidance often leads to more effective optimization.

\begin{figure}[!t]
    \centering

    \begin{subfigure}[t]{0.22\textwidth}
        \centering
        \includegraphics[height=3cm]{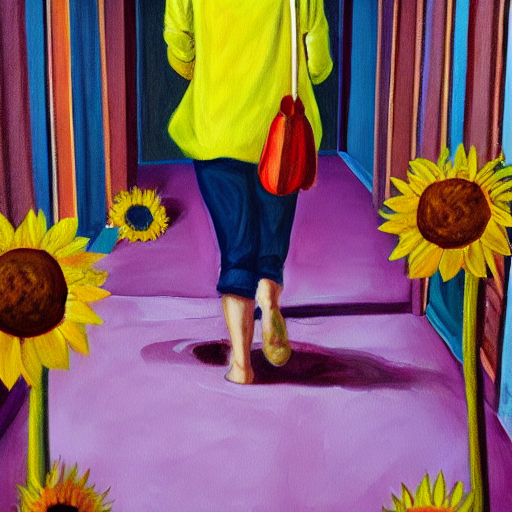}
        \caption{{Unguided} (ImageReward: \num{0.31})}
        \label{fig:hpr guidance 1(a)}
    \end{subfigure}
    \quad
    \begin{subfigure}[t]{0.22\textwidth}
        \centering
        \includegraphics[height=3cm]{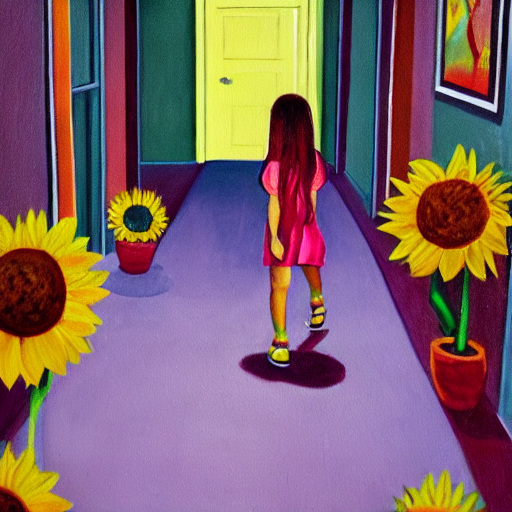}
        \caption{{Forward universal guidance by exact gradients} (ImageReward: \num{1.61})}
        \label{fig:hpr guidance 1(d)}
    \end{subfigure}
    \quad
    \begin{subfigure}[t]{0.22\textwidth}
        \centering
        \includegraphics[height=3cm]{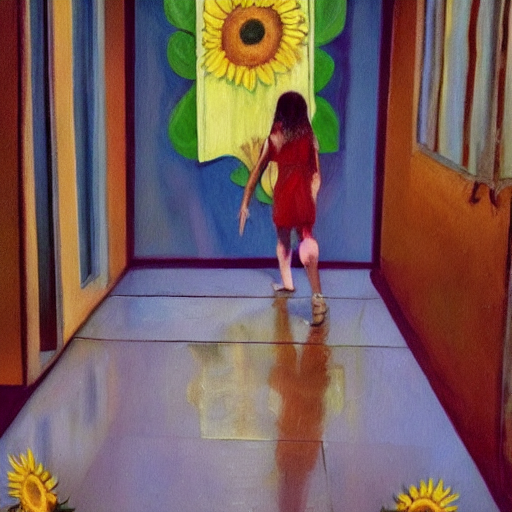}
        \caption{{\subA} \textbf{(Round 21, ImageReward: $\mathbf{1.29}$)}}
        \label{fig:hpr guidance 1(b)}
    \end{subfigure}
    \quad
    \begin{subfigure}[t]{0.22\textwidth}
        \centering
        \includegraphics[height=3cm]{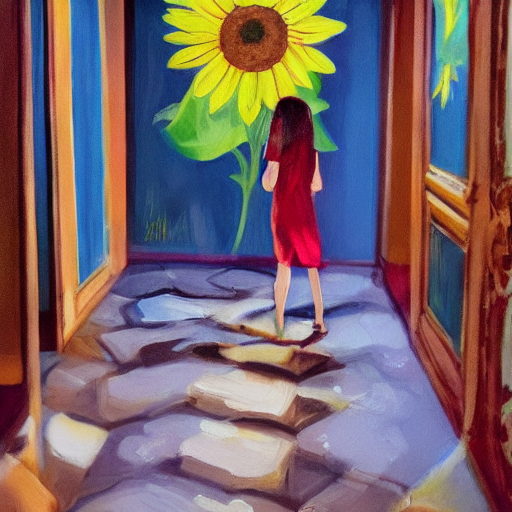}
        \caption{{\subB} \textbf{(Round 11, ImageReward: $\mathbf{1.75}$)}}
        \label{fig:hpr guidance 1(c)}
    \end{subfigure}

    \caption{\textbf{Human preference-guided generation (Task 2, Section~\ref{sec: 4.3}).} \textit{A painting of a girl walking in a hallway and suddenly finds a giant sunflower on the floor blocking her way}. \textsc{BReaD} generates high-score samples with only a few evaluations without accessing reward gradients. Images are generated with the same seed.}
    \label{fig:hpr guidance 1}
\end{figure}

\textbf{Extension to Deep Kernel Learning.}\quad 
To construct expressive kernel functions in high-dimensional spaces, we utilize deep kernel learning~\citep[DKL]{wilson2015deepkernellearning}, which uses a deep neural network to transform data into a low-dimensional representation, i.e., $g(\cdot; w): \gX \to \sR^d$. We then define the GP kernel over the learned representations as
\begin{equation*}
    \kappa_\vw(x, x^\prime) := \kappa (g(x; \vw), g(x^\prime; \vw)),
\end{equation*}
where $\kappa$ is a base kernel function, e.g., the radial basis function (RBF) kernel. The weights can be learned by maximizing the marginal likelihood of the observed data
\begin{equation*}
    \max_{\vw} \left\{ \log p(y_{1:N} | x_{1:N}; \vw) \propto - y_{1:N}^\top \left( K_\vw^N + \tau^2 I \right)^{-1} y_{1:N} - \log \det (K_\vw^N + \tau^2 I ) \right\},
\end{equation*}
where $K_\vw = [\kappa_\vw (x_i, x_j)]$ is the kernel matrix.

%% file: 5-exp.tex


\begin{figure}[!t]
    \centering
    \begin{subfigure}[t]{0.42\textwidth}
        \centering
        \includegraphics[height=4.4cm]{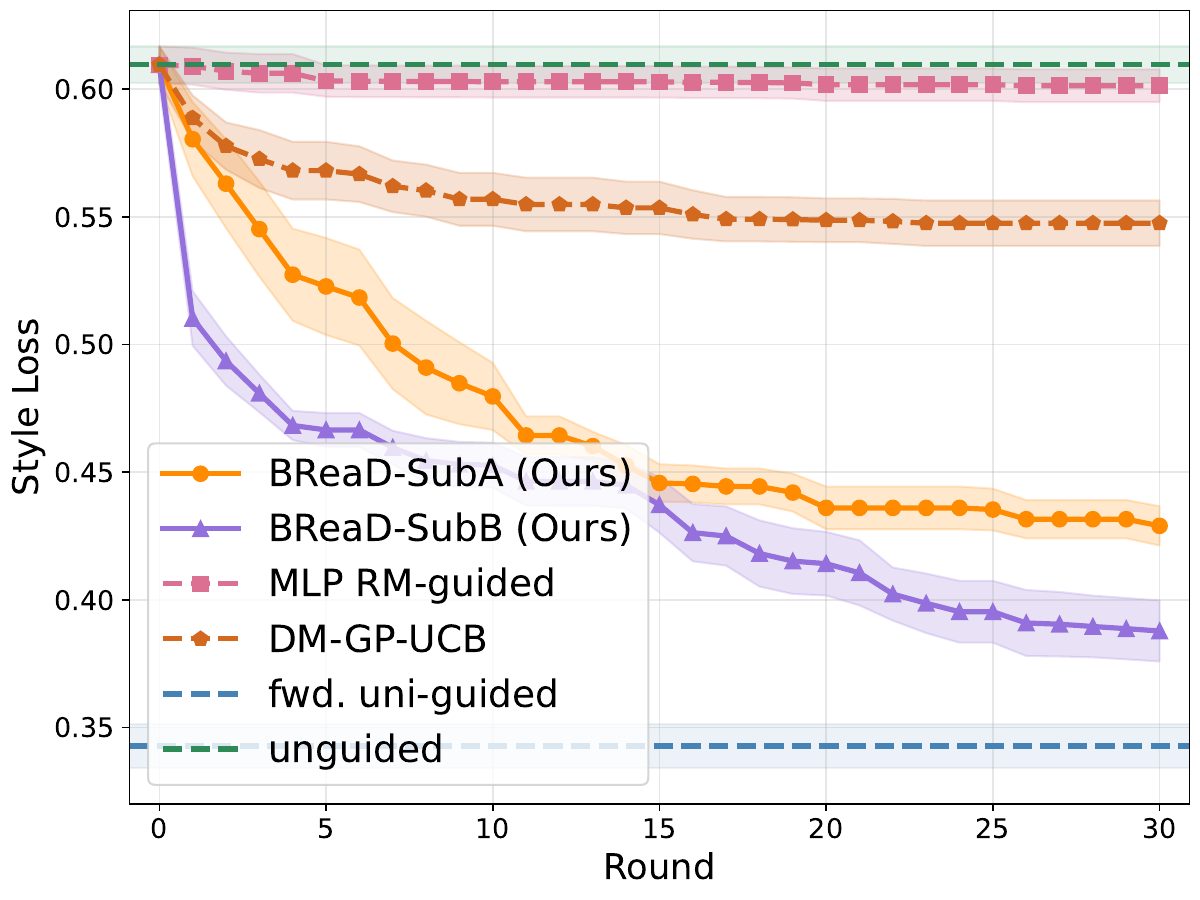}
        \caption{Style guidance (Task 1): Matching the text prompt, \textit{A portrait of a woman}, guided by Pablo Picasso’s painting, \textit{Brick Factory at Tortosa}}
    \label{fig:sg_curve}
    \end{subfigure}
    \qquad
    \begin{subfigure}[t]{0.42\textwidth}
        \centering
        \includegraphics[height=4.4cm]{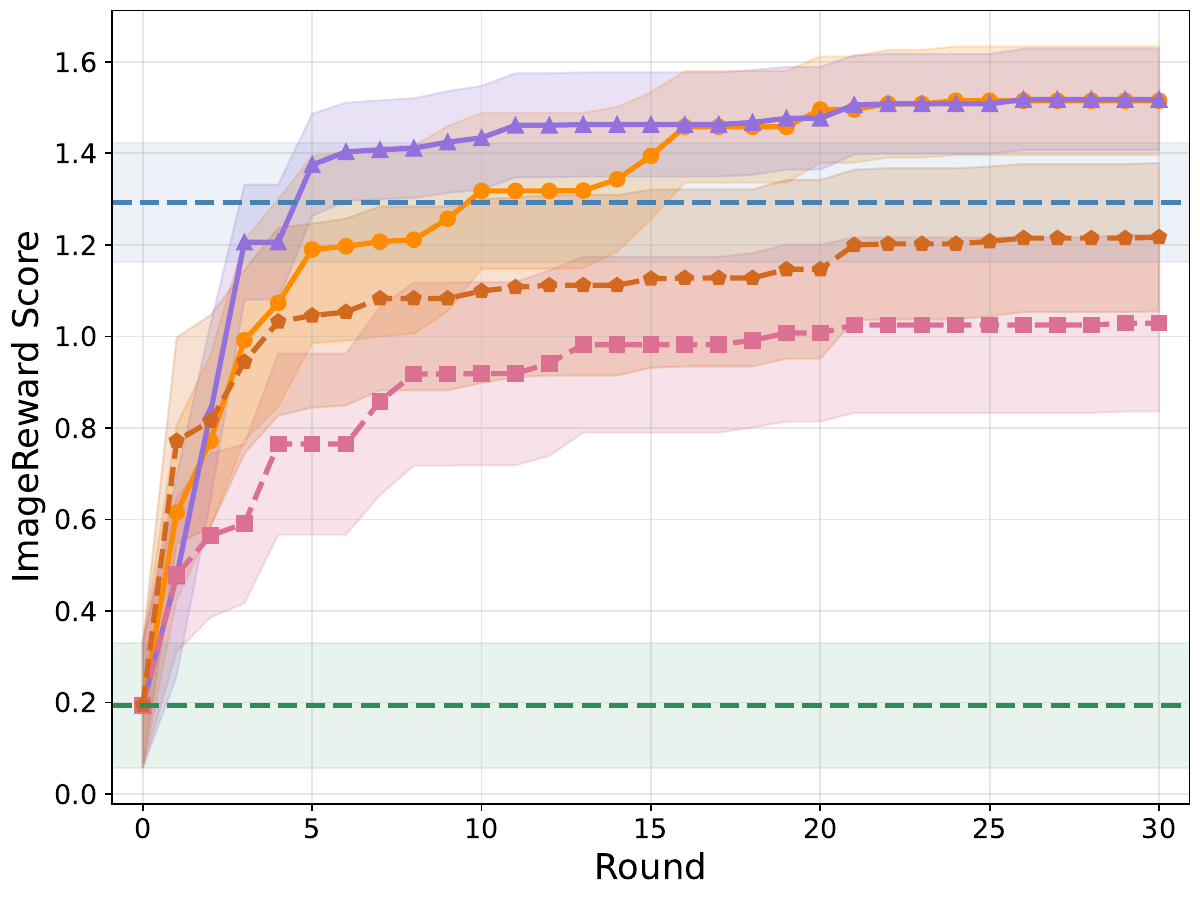}
        \caption{Human-preference guidance (Task 2): \textit{A painting of a girl walking in a hallway and suddenly finds a giant sunflower on the floor blocking her way}}
    \label{fig:hp_curve}
    \end{subfigure}

    \caption{\textbf{Image generation guided by black-box rewards (Section~\ref{sec: 4.3}).} Optimization for style loss~(Left) and ImageReward score~(Right). Results are averaged over 10 trials.}
    \label{fig:style guidance_main text}
\end{figure}

\textbf{Baselines.}\quad 
In this section, we report the performance of the proposed framework across image generation, protein design, and a simulated nuclear fusion control task. Particularly, we compare \subA~(\textsc{BReaD} with \textsc{Subroutine A}) and \subB~(\textsc{BReaD} with \textsc{Subroutine B}) against the following baselines: 1) \mlp, which guides the generation through a three-layer multi-layer perceptron reward model trained at each iteration to predict the black-box objective using all observations collected so far; 2) \gpucb, GP-UCB~\citep{3104322.3104451} using the diffusion model~(DM) as a priori for searching; it generates \num{10} candidate samples from the DM at each iteration and selects the one with the highest UCB value for evaluation; 3) \fwd, which applies Forward universal guidance~\citep{bansal2024universal} using the exact reward gradients of the black-box objective; 4) \unguided, which generates samples from the pre-trained diffusion model without reward guidance (while retaining prompt guidance in image generation)). We use the labels for \num{20} unconditionally generated samples to initialize the GP in \textsc{BRead} and DM-GP-UCB methods and the reward model in the MLP RM-guided baseline. We report both the mean and the standard error. When reporting the results, we plot the curve for the best-so-far observed loss/reward for BReaD-SubA, BReaD-SubB, MLP-guided, GP-UCB, and random search. Since the unguided and fwd. uni-guided baselines do not adapt to new observations, we report the best loss/reward among \num{30} generated images.

\textbf{Summary of results.}\quad 
We summarize the numerical results as follows. The proposed methods achieve performance comparable to the fwd. uni-guided baseline, despite not having access to the exact reward gradients. Meanwhile, they also consistently outperform the MLP-guided baseline. This demonstrates the benefit of GP-based surrogate modeling and guidance in low-data settings. Compared with GP-UCB, which follows a generate-then-select strategy over candidates sampled unconditionally from the diffusion model, our methods consistently generate higher-reward samples, especially when such samples may have low probability under the unconditional data distribution. Finally, incorporating UCB into the guidance process further promotes exploration of promising regions and often leads to more effective optimization performance. Additional experimental results and implementation details are postponed to Appendix~\ref{sec: additional results}.

\subsection{Results for Image Generation}
\label{sec: 4.3}

\textbf{Task 1. Style-guided generation.}\quad
We perform style-guided image generation using Stable Diffusion v1.5~\citep{9878449}. The \textit{style loss} is defined as the cosine distance between the CLIP embeddings~\citep{pmlr-v139-radford21a} of the generated and reference images, i.e., $\ell_\textup{style}(x_\text{gen}) = 1 - \cos\langle CLIP(x_\text{ref}), CLIP(x_\text{gen})\rangle$, where $x_\text{ref}$ is the image in the reference style and $x_\text{gen}$ is the generated image. Figure~(\ref{fig:sg_curve}) reports results for generating images that match the text prompt, \textit{A portrait of a woman}, and the style of Pablo Picasso’s painting, \textit{Brick Factory at Tortosa}. Example generated images can be found in Figure~\ref{fig:style guidance 3} in the Appendix. Our methods can effectively produce images that match the reference style and achieve a loss comparable to the fwd. uni-guided baseline that has access to the exact loss gradient. Moreover, incorporating UCB into guidance leads to faster reduction in style loss.

\textbf{Task 2. Human preference-guided generation.}\quad
One practical low-data setting is personalized image generation, where a user may wish to generate images with desired properties based on only a few rounds of feedback. To simulate such feedback, we use ImageReward~\citep{xu2023imagereward}, a text-to-image human-preference reward model trained on \num{137}k expert comparisons, as the black-box objective. We generate images with Stable Diffusion v1.5 and optimize the ImageReward score through iterative queries. Figures~(\ref{fig:hp_curve}) and~\ref{fig:hpr guidance 1} show one such example where \textsc{BReaD} effectively discovers images with substantially higher rewards and fidelity using only a small number of evaluations.

\subsection{Results for Ligand optimization}
\label{sec: ligand}

\begin{figure}[!t]
    \centering
    \begin{subfigure}[t]{0.42\textwidth}
        \centering
        \includegraphics[height=4.4cm]{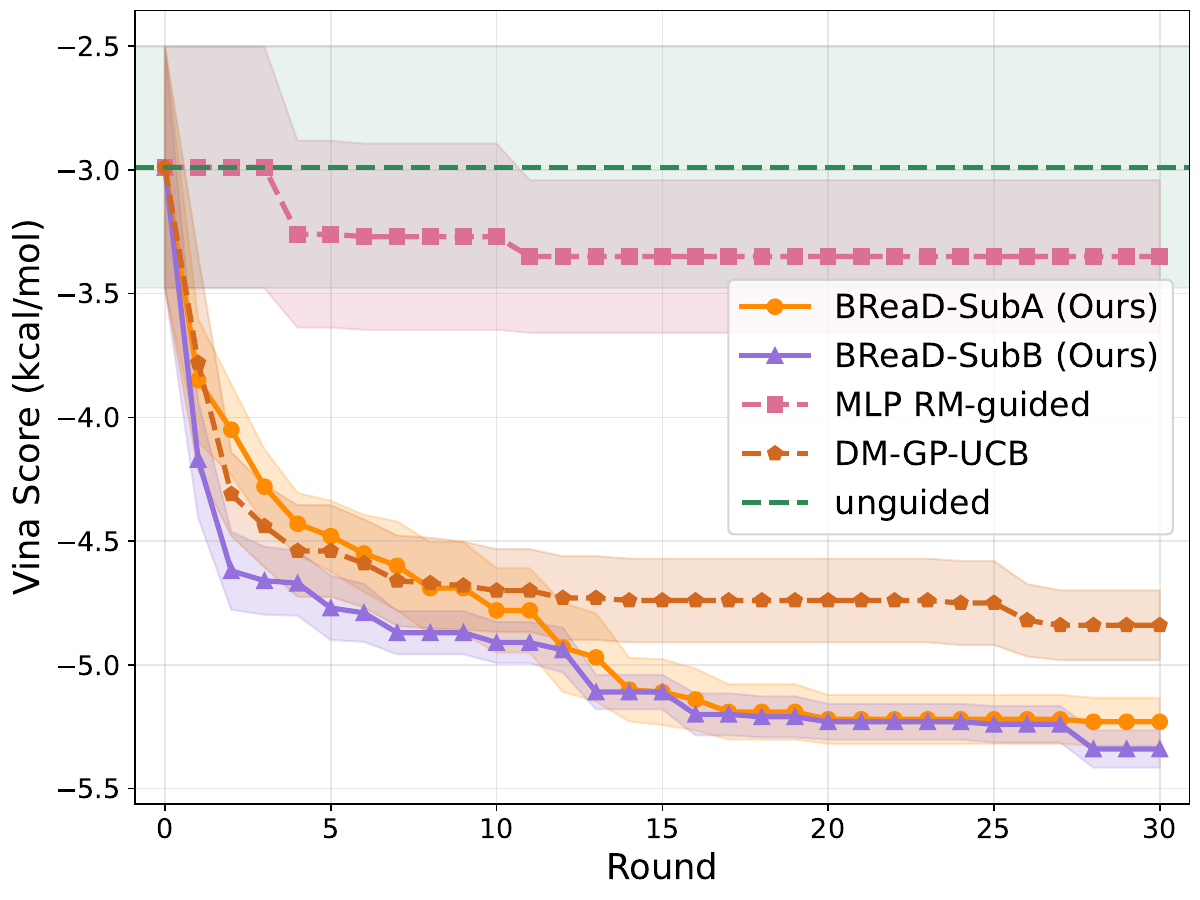}
        \caption{Dipeptidyl peptidase-4 (DPP4)}
        \label{fig:prot_desgn_1}
    \end{subfigure}
    \quad
    \begin{subfigure}[t]{0.42\textwidth}
        \centering
        \includegraphics[height=4.4cm]{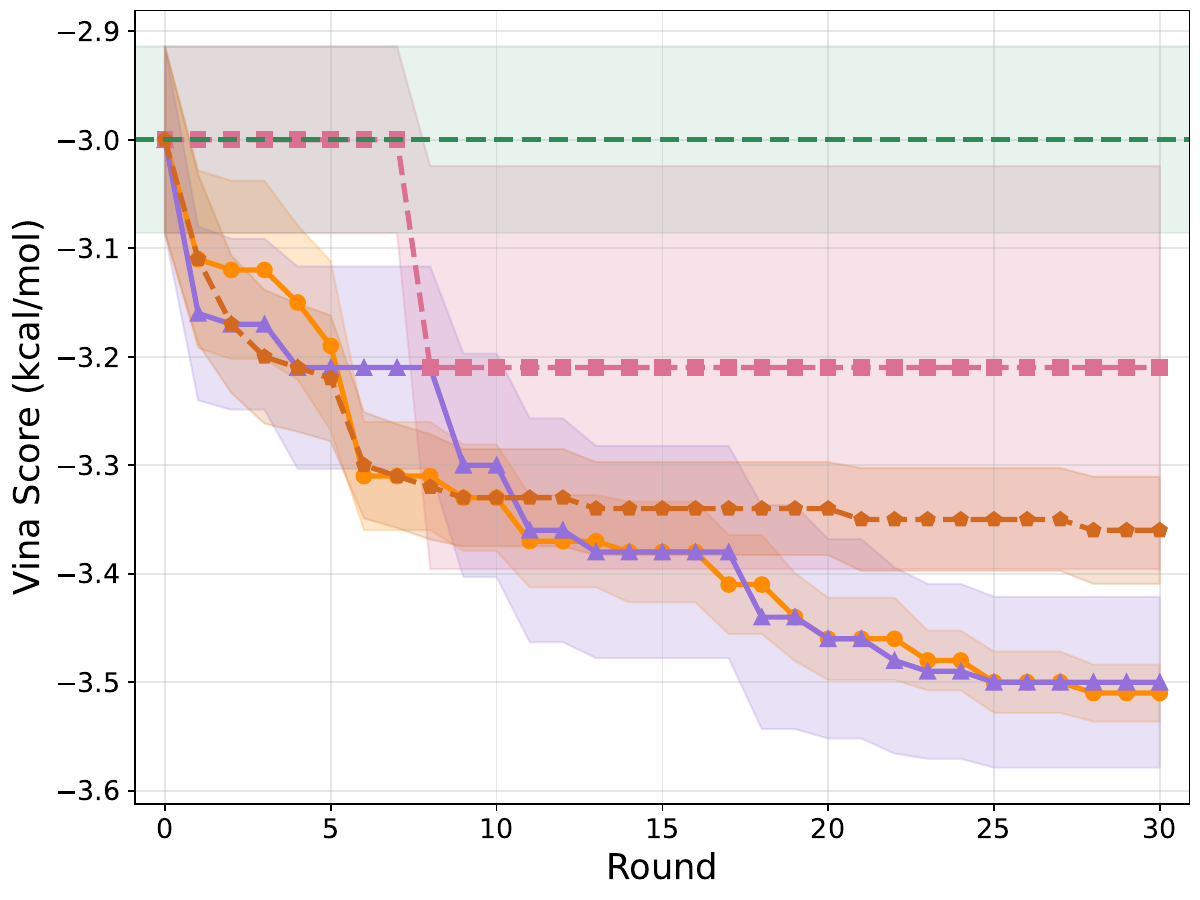}
        \caption{p38 MAP kinase (MAPK)}
        \label{fig:prot_desgn_2}
    \end{subfigure}
    \caption{\textbf{Ligand optimization for Vina scores.} DPP4 and p38-MAPK. Results are averaged over \num{10} trials.}
    \label{fig:prot_desgn}
\end{figure}

Structure-based drug design aims to identify ligand molecules that bind strongly to a target protein. We consider optimizing the Vina score~\citep{https://doi.org/10.1002/jcc.21334}, a widely used molecular docking score that approximates the binding affinity between a ligand and a protein binding pocket. A lower Vina score indicates favorable predicted binding. Directly applying gradient-based optimization methods can be challenging since the Vina score is non-differentiable. We use the pre-trained DiffSBDD model\footnote{Github repository: \url{https://github.com/arneschneuing/DiffSBDD}. We use the conditional full-atom model in the experiment.}~\citep{Schneuing2024}, which generates three-dimensional ligand structures conditioned on a target protein pocket. Figure~\ref{fig:prot_desgn} reports results on DPP4 and p38-MAPK. We do not include the forward universal-guidance baseline because gradients of the Vina scoring function are unavailable. \textsc{BReaD} effectively improves the Vina score for both protein targets using only a small number of evaluations and outperforms the baselines. 

\subsection{Results for Controllable Nuclear Fusion}
\label{sec: wpqh exp}

Nuclear fusion is a promising path toward sustainable energy. Our goal is to access the Wide-Pedestal QH mode~\citep[WPQH]{Li_2022}, an advanced plasma scenario that achieves high energy confinement without damaging transient heat pulses. The labeled data for this plasma scenario is scarce. Among approximately \num{10}k shots in the historical dataset of the DIII-D Tokamak in San Diego, we only have the WPQH labels for around \num{400} shots. We pre-trained a diffusion model on the historical shot data. Specifically, given the plasma state and actuator sequence during the first \num{200}ms of the flat-top phase (\num{10} control steps), the diffusion model models the plasma state and actuator sequence over the subsequent \num{3800}ms (\num{190} control steps). In the experiment, we parameterize six controllable actuators, including beam power, beam torque, $I_p$, $q95$, and ECH, using 21 parameters that determine their control waveforms.

\textbf{Simulation results.}\quad
We evaluate the methods in simulation. To simulate the tokamak dynamics, we train an ensemble of \num{25} recurrent probabilistic neural networks (RPNNs) on the historical DIII-D dataset, following~\citep{char2024shotpredictionsdiiidtokamak, pmlr-v267-sonker25a}. Given an actuator configuration, the RPNN ensemble predicts the resulting plasma trajectories. Additionally, we train a WPQH detector using labeled experimental data to identify whether the plasma is in WPQH mode at each timestep. The trained detector achieves \num{85.49}\% accuracy on real experimental trajectories and \num{72.46}\% accuracy when applied to trajectories rolled out by the RPNN ensemble. We then compute the WPQH ratio as the fraction of time steps during the flat-top phase classified as WPQH. We provide additional details in Appendix~\ref{sec: C}. Figure~\ref{fig:wpqh} reports the best-so-far WPQH ratio over 30 optimization rounds. The results show that the proposed uncertainty-aware guidance method progressively improves the WPQH ratio and matches the performance of the forward universal guidance baseline that has direct access to the objective gradients.

\begin{figure}[!h]
    \centering
    \includegraphics[width=0.5\linewidth]{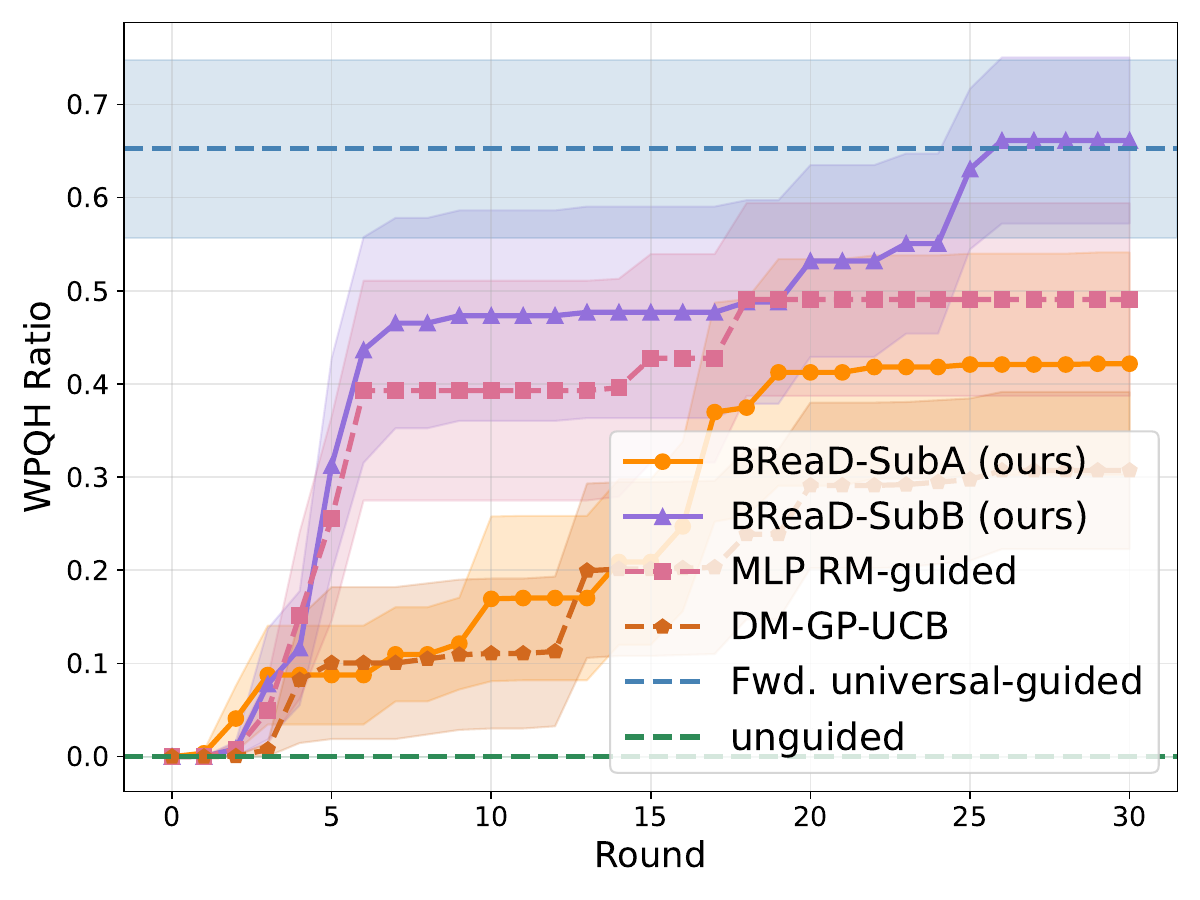}
    \caption{Accessing the Wide-Pedesatal QH in a simulated tokamak environment. Results are averaged over \num{10} trials.}
    \label{fig:wpqh}
\end{figure}

%% file: 6-discussion.tex
We propose \textsc{BReaD}, a framework for optimizing black-box reward functions in high-dimensional data spaces. \textsc{BReaD} uses a Gaussian process as a differentiable surrogate for the unknown reward function and constructs guidance signals for guided generation from a diffusion model that is pre-trained on unlabeled data. Numerical results on image generation, protein design, and simulated fusion plasma control tasks demonstrate that our framework can effectively optimize black-box rewards in low-data scenarios. An important direction for future work is to deploy and evaluate \textsc{BReaD} on real-world systems, such as the DIII-D tokamak. Another promising direction is to extend the framework to multi-objective and constrained optimization problems, which commonly arise in scientific applications.

%% file: appendix.tex
\section{Related Work}
\label{2-related_work}
\input{2-related_work}

\section{Distibution Mismatch Analysis of Subroutine A}
\label{sec: B}

\subsection{Notations and Definitions}

\begin{longtable}{c c}
    \label{tab:notation} \\

    \toprule
    \textbf{Notation} & \textbf{Definition} \\
    \midrule
    \endfirsthead

    \toprule
    \textbf{Notation} & \textbf{Definition} \\
    \midrule
    \endhead

    \midrule
    \multicolumn{2}{r}{\textit{Continued on next page}} \\
    \endfoot

    \bottomrule
    \endlastfoot

    $\fstar$ & (Unknown) Reward function \\
    $\gN(0,\tau)$ &  Noise in the observed rewards \\
    $\ystar$ & Target score \\

    \midrule

    $\qd$ & Data distribution \\     
    $q_t$ & Marginal distribution of $\bar X_t$ in process~(\ref{eq: qt}) \\
    $Q_T^\leftarrow$ & Path distribution of the backward SDE~(\ref{eq: rev_Q}) \\

    \midrule

    $m_t(x_t)$ & Posterior mean $\E_{q_0(\cdot|x_t)}[x_0]$~(\ref{eq: posterior_mean}) \\

    \midrule

    $J_m(x_t, t)$ &
    \makecell[c]{Jacobian matrix $\nabla_{x_t} m_t(x_t)$ \\
    of the posterior-mean map} \\

    \midrule

    $q_\ystar (x) \propto \qd(x) \gN(\ystar;\fstar(x),\tau^2)$
    & Target-conditioned distribution \\

    $Z_\ystar$
    & Normalizer $\int \gN(\ystar; \fstar(x_0), \tau^2)
    \qd(x_0) \dd x_0$ \\

    $q_{\ystar,t}$
    & Marginal distribution of $\bar X_{\ystar,t}$
    in process~(\ref{eq: q_yt}) \\

    $Q_{\ystar, T}^\leftarrow$
    & Path distribution of the backward SDE~(\ref{eq: rev_Qystar}) \\

    $\bar p_t(\ystar|x_t)$
    & $\E_{x_0 \sim q_0(\cdot|x_t)}
    [\gN(\ystar; \fstar(x_0), \tau^2)]$ \\

    \midrule

    $\qGpk (x) \propto \qd(x)\gN(\ystar;\mu_k(x),\sigma_k^2(x)+\tau^2)$
    & \makecell[c]{GP-conditioned distribution \\
    at the $k$-th iteration} \\

    \midrule

    $Z_k$
    & Normalizer $\int \qd(x)
    \gN(\ystar;\mu_k(x),\sigma_k^2(x)+\tau^2) \dd x$ \\

    $\qGpkt$
    & Marginal distribution of $\barX_t^k$
    in process~(\ref{eq: q_GP_yt}) \\

    $Q_{\textup{GP},\ystar,T}^{k,\leftarrow}$
    & Path distribution of the backward SDE~(\ref{eq: rev_GP}) \\

    $\bar p^k_{\textup{GP}, t}(\ystar|x_t)$
    & $\E_{x_0 \sim q_0(\cdot|x_t)}
    [\gN(\ystar; \hf^k(x_0), \tau^2)]$ \\

    \midrule

    $\hq^{k}_\ystar$
    & \makecell[c]{Generated distribution at the \\
    $k$-th online iteration} \\

    \midrule

    $\hf^{k}$
    & \makecell[c]{GP posterior mean of $\fstar$ fitted \\
    at the $k$-th online iteration} \\

    \midrule

    $\hat{x}_0^\theta(x_t, t)$
    & Predicted posterior mean~(\ref{eq: predicted_mean}) \\

    $J_{\hat x_0^\theta}(x_t, t)$
    & \makecell[c]{Jacobian matrix
    $\nabla_{x_t} \hat x_0^\theta (x_t, t)$ of the \\
    posterior-mean prediction map} \\

    \midrule

    $\epsilon_\textup{uncond}^2$
    & \makecell[c]{Bound on the per-step score \\
    estimation error~(\ref{eq: error_uncond})} \\

    \midrule

    $\bar\epsilon_\textup{uncond}^2$
    & \makecell[c]{Bound on the weighted-average score \\
    estimation error~(\ref{eq: error_bar_uncond})} \\

    \midrule

    $\bar\epsilon_\textup{Hess}^2$
    & \makecell[c]{Bound on the weighted-average \\
    score Hessian estimation error~(\ref{eq: error_hess})} \\

    \midrule

    $\epsilon_{\textup{post-avg}}^2$
    & Posterior-averaging error~(\ref{eq: error_post_avg}) \\

    $L_\tau$
    & (\ref{eq: Ltau}) \\

    $L_{\tau,\var}$ & (\ref{eq: Ltau_var}) \\

    $h$ & Discretization length \\

    $\mathfrak{m}_2^2$
    & Assumption~\ref{as: second_moment} \\

    $G_k$ & Assumption~\ref{as: der_log_gp} \\
    $\hat J_{\max}$ & (\ref{eq: constant_hJ}) \\
    $J_{\max}$ & (\ref{eq: constant_J}) \\
    $B^\star$ & Assumption~\ref{as: bounded_reward} \\
    $r_k$ & GP mean-approximation error (\ref{eq: constant_rk}) \\
    $r_{\nabla, k}$ & GP gradient-approximation error (\ref{eq: constant_rnk}) \\
    $v_k^2$ & GP variance~(\ref{eq: gp-var}) \\
    $v_{\nabla,k}$ &  Gradient of the GP variance~(\ref{eq: gp-var-grad}) \\
    $M_k$ & (\ref{eq: M_k}) \\
    $B_k$ & (\ref{eq: B_k}) \\
    $c_\tau$ & (\ref{eq: constant_ctau}) \\
    $c_{\tau,k}$ & (\ref{eq: constant_ctauk}) \\
    $L_k$ & Assumption~\ref{as: lagrange_score} \\
    $\tilde{L}_k$ & Assumption~\ref{as: lagrange_score} \\
\caption{Notation and definitions in the proof}
\end{longtable}

\subsection{Preliminaries}
\label{sec: B.2}

\paragraph{Unconditional generation processes}

In the proof, we consider $g(t) \equiv 1$ in the forward SDE~(\ref{eq: forward_process}), that is,
\begin{equation}
    \dd \barX_t = - \barX_t \dd t + \sqrt{2} \dd B_t, X_0 \sim \qd,
\label{eq: qt}
\end{equation}
and we have $q_t(x_t|x_0) = \gN(x_t; \sqrt{\bar \alpha_t} x_0, 1 - \bar \alpha_t I)$, where $\bar \alpha_t = \exp(-2t)$. The corresponding backward SDE is given by
\begin{equation}
    \dd \barXr_t = \left\{ \barXr_t + 2 \nabla \log q_{T-t} (\barXr_t) \right\} \dd t + \sqrt{2} \dd B_t, \barXr_0 \sim q_{T}.
\label{eq: rev_Q}
\end{equation}

\paragraph{Sampling from the target distribution}

Next, we consider the processes to sample from the target distribution, i.e., $q_\ystar(x) \propto \qd(x) \gN(\ystar; \fstar(x), \tau^2)$. The forward process is given by
\begin{equation}
    \dd \barX_{\ystar,t} = - \barX_{\ystar,t} \dd t + \sqrt{2} \dd B_t, X_0 \sim q_\ystar.
\label{eq: q_yt}
\end{equation}
The corresponding backward SDE is given by
\begin{equation}
    \dd \barXr_{\ystar,t} = \left\{ \barXr_{\ystar,t} + 2 \nabla \log q_{\ystar, T-t} (\barXr_{\ystar,t}) \right\} \dd t + \sqrt{2} \dd B_t, \barXr_{\ystar,0} \sim q_{\ystar, T}.
\label{eq: rev_Qystar}
\end{equation}
Specifically, we have
\begin{align*}
    \nabla \log q_{\ystar, T-t} (\barXr_{\ystar,t}) = \nabla \left[ \log q_{T-t} (\barXr_{\ystar,t}) + \log \bar p_{T-t}(\ystar | \barXr_{\ystar,t}) \right],
\end{align*}
where $\bar p_{T-t}(\ystar | \barXr_{\ystar,t}) = \E_{x_0 \sim q_0(\cdot|x_t)} [\gN(\ystar; \fstar(x_0), \tau^2)]$. To show this, note that
\begin{equation}
    q_\ystar(x_0) = \frac{\qd(x_0) \gN(\ystar; \fstar(x_0), \tau^2)}{Z_\ystar},
\end{equation}
where $Z_\ystar = \int \gN(\ystar; \fstar(x_0), \tau^2) \qd(x_0) \dd x_0$ is the normalizer. Therefore, we have
\begin{align*}
    q_{\ystar, t}(x_t) = & \int q_t(x_t|x_0) q_\ystar (x_0) \dd x_0 = \frac{1}{Z_\ystar} \int q_t(x_t|x_0) \qd(x_0) \gN(\ystar; \fstar(x_0), \tau^2) \dd x_0 \\
    = & \frac{q_t(x_t)}{Z_\ystar} \int \frac{q_t(x_t|x_0) \qd (x_0)}{q_t(x_t)} \gN(\ystar; \fstar(x_0), \tau^2) \dd x_0 \\
    = & \frac{q_t(x_t)}{Z_\ystar} \E_{x_0 \sim q_0(\cdot|x_t)} [\gN(\ystar; \fstar(x_0), \tau^2)].
\end{align*}
Taking the logarithm and the derivative on both sides leads to the statement.

\paragraph{(Ideal) Sample distribution}

Since we model the unknown reward function with a GP, we consider the processes to sample from $\qGpk(x) \propto \qd(x) \gN (\ystar; \mu_k(x), \sigma_k^2(x)+\tau^2)$.
\begin{equation}
    \dd \barX_t^k = - \barX_t^k \dd t + \sqrt{2} \dd B_t, X_0^k \sim \qGpk.
\label{eq: q_GP_yt}
\end{equation}
The corresponding backward SDE transforms noise into samples from $\qGpk$
\begin{equation}
    \dd \barXrk_t = \left\{ \barXrk_t + 2 \nabla \log \qGpktr (\barXrk_t) \right\} \dd t + \sqrt{2} \dd B_t, \barXrk_0 \sim \qGpkT.
\label{eq: rev_GP}
\end{equation}
Specifically, we have
\begin{align*}
    \nabla \log \qGpktr (\barXrk_t) = \nabla \left[ \log q_{T-t}(\barXrk_t) + \log \bar p^k_{\textup{GP}, T-t} (\ystar | \barXrk_t) \right],
\end{align*}
where $\bar p^k_{\textup{GP}, T-t} (\ystar | \barXrk_t) = \E_{x_0 \sim q_0(\cdot|x_t)} [\gN(\ystar; \mu_k(x_0), \sigma_k^2(x_0)+\tau^2)]$.

Our proof is based on the following assumptions.
\begin{assumption}[Bounded reward]
We have $\fstar(x) \in [0,1]$ and $\| \nabla \fstar(x) \| \le B^\star$ for any $x \in \gX$.
\label{as: bounded_reward}
\end{assumption}

\begin{assumption}
\label{as: f}
The true reward $\fstar: \gX \to [0,1]$ belongs to the RKHS $\gH_\kappa$ associated with the kernel $\kappa(\vx, \vx^\prime)$ and satisfies $\|\fstar\|_{\gH_\kappa} \le B$. At online iteration $k$, the observed reward is $y^k = \fstar(x^k) + \varepsilon^k$, where the noise $\varepsilon^k$ is conditionally zero-mean given the history and is $R$-sub-Gaussian.
\end{assumption}

\begin{assumption}[Unconditional score estimation error]
We have
\begin{equation}
    \E_{x_t \sim q_{t}} \left[ \left\| s_t^\theta(x_t) - \nabla \log q_t (x_t) \right\|^2 \right] \le \epsUncond^2
\label{eq: error_uncond}
\end{equation}
\label{as: score_error}
\end{assumption}

\begin{assumption}[Bounded kernel]
We have $\sup_{x \in \gX} \kappa(x,x) \le \kappa_{\max}$.
\end{assumption}

\begin{assumption}[Adapt from {\cite[A1]{chen2023sampling}}]
The score function $\nabla \log \qGpkt$ of the forward process~(\ref{eq: q_GP_yt}) is $L_k$-Lipschitz, i.e., $\| \nabla \log \qGpkt(x) - \nabla \log \qGpkt(x') \| \le L_k \| x - x' \|$.
\label{as: lagrange_score}
\end{assumption}

\begin{assumption}[Bounded second moment {\cite[A2]{chen2023sampling}}]
The second moment of the data distribution $\qd$ is finite, i.e., $\mathfrak{m}_2^2 := \E_\qd[\|\cdot\|^2] < \infty$.
\label{as: second_moment}
\end{assumption}

\begin{assumption}[{\cite[A3]{chen2023sampling}}]
    The data distribution $\qd$ has finite KL divergence w.r.t. the standard Gaussian.
\label{as: finite_KL}
\end{assumption}

\begin{assumption}
    $\nabla \log p_{\textup{GP}}^k(\ystar|x)$ is bounded and $\tilde{L}_k$-Lipschitz, i.e., $\| \nabla \log p_{\text{GP}}^k (\ystar | x ) \| \le G_k$ and $\| \nabla \log p_{\textup{GP}}^k(\ystar|x) - \nabla \log p_{\textup{GP}}^k(\ystar|x') \| \le \tilde{L}_k \| x - x' \|$ for any $x,x' \in \gX$.
\label{as: der_log_gp}
\end{assumption}

\subsection{Complete Distribution Mismatch Bound}
\label{sec: B.3}

We state the complete bound in Theorem~\ref{thm: main}.
\begin{equation}
\begin{aligned}
    & \TV(q_\ystar, \hq^k_{\ystar}) \\
    \lesssim & \underbrace{ \frac{L_\tau r_k + L_{\tau,\var} v_k^2 }{c_\tau} \left( 1 + \sqrt{B_k + \frac{M_k^2}{c_{\tau,k}^2} T} \right) + \frac{ J_{\max} }{\tau^2}  \left[ (1 + r_k) r_{\nabla, k} + \left( r_k + \frac{v_k^2}{\tau^2} \right) B^\star + \frac{v_{\nabla,k}}{2} \left( \frac{ (1 + r_k)^2 }{ \tau^2 } + 1 \right) \right] \sqrt{T} }_{\varepsilon_{\textup{GP}, k}\textup{ (GP approximation error)}} \\
    & + \underbrace{ \left( 1 + \frac{v_k^2}{\tau^2} \right)^{1/4} \exp\left( \frac{(1 + r_k)^2 + 2}{4\tau^2} \right) \sqrt{ \epsilon_\textup{uncond}^2 T + \hat{J}_{\max}^2 \tilde{L}_k^2 \bar \epsilon_{\textup{uncond}}^2 + G_k^2 \bar \epsilon_{\textup{Hess}}^2 } }_{\varepsilon_{\textup{score}} \textup{ (score estimation error)}} \\
    & + \underbrace{ L_k\sqrt{dh} + \left( 1 + \frac{v_k^2}{\tau^2} \right)^{\frac{1}{4}} \exp\left( \frac{(1 + r_k)^2}{4\tau^2} \right) L_k h \mathfrak{m}_2 }_{ \varepsilon_{\textup{disc}} \textup{ (discretization error)}} \\
    & + \underbrace{ \left( 1 + \frac{v_k^2}{\tau^2} \right)^{\frac{1}{4}} \exp\left( \frac{(1 + r_k)^2 + 1}{4\tau^2} \right) \epsilon_\textup{post-avg} }_{\varepsilon_{\textup{post-avg}} \textup{ (posterior-averaging error)}} \\
    & + \underbrace{ \sqrt{\KL(\qGpk \| \varepsilon)} \exp(-T) }_{ \varepsilon_{\textup{init}} \textup{ (Initialization error)}},
\end{aligned}
\label{eq: distribution mismatch}
\end{equation}

\subsection{Proof of Theorem~\ref{thm: main}}

To this end, we first decompose the suboptimality of $\hq^{k}_\ystar$, i.e., the generated distribution at the $k$-th iteration. Note that
\begin{align}
    \TV(q_\ystar, \hq^k_{\ystar}) \le \underbrace{\TV(q_\ystar, \qGpk)}_{\gE_1\textup{ (GP approximation error)}} + \underbrace{\TV(\qGpk, \hq^k_{\ystar})}_{\gE_2 \textup{ (Sampling error)}},
\end{align}
where $\qGpk(x) \propto \qd(x) \gN(\ystar; \mu_k(x), \sigma^2_k(x) + \tau^2)$ is the target distribution induced by the GP.

\subsubsection{Bouding GP Approximation Error \texorpdfstring{$\gE_1$}{E1}}
\label{sec: b.3.1}
Assume
\begin{gather}
    \quad \sup_x |\mu_k(x) - \fstar(x)| \le r_k, \label{eq: constant_rk} \\
    \sup_x \sigma_k^2(x) \le v_k^2.
\label{eq: gp-var}
\end{gather}
We show that
\begin{equation*}
    \gE_1 = \TV(q_\ystar, \qGpk) \le \frac{L_\tau r_k + L_{\tau,\var} v_k^2 }{c_\tau}.
\end{equation*}
Note that by definition of total variation distance, we have
\begin{align*}
    \TV(q_\ystar, \qGpk) = & \frac{1}{2} \int \qd(x) \left| \frac{ \gN(\ystar; \fstar(x), \tau^2)}{Z_\ystar} - \frac{\gN(\ystar; \mu_k(x),\sigma_k^2(x)+\tau^2)}{Z_k} \right| \dd x \\
    \le & \frac{1}{2} \Bigg(\int \qd(x) \left| \frac{ \gN(\ystar; \fstar(x), \tau^2) - \gN(\ystar; \mu_k(x),\sigma_k^2(x)+\tau^2)}{Z_\ystar} \right| \dd x \\
    & + \left| \frac{1}{Z_\ystar} - \frac{1}{Z_k} \right| \underbrace{\int \qd(x) \gN(\ystar; \mu_k(x),\sigma_k^2(x)+\tau^2) \dd x}_{=Z_k} \Bigg)  \\
    \le & \frac{1}{2} \frac{2}{Z_\ystar} \int \qd(x) \left| \gN(\ystar; \fstar(x), \tau^2) - \gN(\ystar; \mu_k(x),\sigma_k^2(x)+\tau^2) \right| \dd x \\
    \le & \frac{1}{Z_\ystar} \E_{\qd} \left[\left|\gN(\ystar; \fstar(x), \tau^2) - \gN(\ystar; \mu_k(x),\tau^2) \right|\right]\\
    & + \frac{1}{Z_\ystar} \E_{\qd} \left[\left| \gN(\ystar; \mu_k(x),\tau^2) - \gN(\ystar; \mu_k(x),\sigma_k^2(x)+\tau^2) \right|\right],
\end{align*}
where the second inequality uses the fact that $|Z_k - Z_\ystar| \le \E_{\qd}[| \gN(\ystar; \fstar(x), \tau^2) - \gN(\ystar; \mu_k(x),\sigma_k^2(x)+\tau^2) |]$. By the Lipschitz property of the Gaussian likelihood in its mean, there exists a constant 
\begin{equation}
    L_\tau := \frac{1}{\sqrt{2\pi e} \tau^2} \ge \sup_{\mu \in \sR} \left| \frac{\partial}{\partial \mu} \gN(\ystar; \mu, \tau^2) \right|,
\label{eq: Ltau}
\end{equation}
such that $|\gN(\ystar; \fstar(x), \tau^2) - \gN(\ystar; \mu_k(x), \tau^2)| \le L_\tau |\mu_k(x) - \fstar(x)| \le L_\tau r_k$. On the other hand, since
\begin{align}
    & \sup_{\mu\in\sR, v \ge \tau^2} \left| \frac{\partial}{\partial v} \gN(\ystar; \mu, v) \right| \nonumber \\
    = & \sup_{\mu\in\sR, v \ge \tau^2} \left| \frac{1}{2 \sqrt{2\pi} v^{3/2}} \exp \left( -\frac{(\ystar - \mu)^2}{2v} \right) \left( \frac{ ( \ystar - \mu )^2}{v} - 1 \right) \right| \nonumber \\
    \le & \frac{1}{2\sqrt{2\pi} \tau^3} =: L_{\tau,\var}, \label{eq: Ltau_var}
\end{align}
where the last inequality uses the fact that $e^{-z/2}|z-1| \le 1$ for $z \ge 0$. Hence, we have $| \gN(\ystar; \mu_k(x),\tau^2) - \gN(\ystar; \mu_k(x),\sigma_k^2(x)+\tau^2) | \le L_{\tau,\var} v_k^2 $. Furthermore, since $\ystar, \fstar(x) \in [0,1]$ by Assumption~\ref{as: bounded_reward}, we have
\begin{equation}
    Z_\ystar = \int \gN(\ystar; \fstar(x), \tau^2 ) \qd (x) \dd x \ge \frac{1}{\sqrt{2\pi} \tau} \exp\left( -\frac{1}{2\tau^2} \right) =: c_\tau,
\label{eq: constant_ctau}
\end{equation}
which concludes the proof.

\subsubsection{Bounding Sampling Error \texorpdfstring{$\gE_2$}{E2}}

First, note that
\begin{align*}
    \gE_2 = \TV(\qGpk, \hq^k_{\ystar}) \le & \TV(\qGpk, \tq^k_{\ystar}) + \TV(\tq^k_{\ystar}, \hq^k_{\ystar}) \\
    \le & \TV(\qGpk, \tq^k_{\ystar}) + \TV(\qGpkT, \varepsilon),
\end{align*}
where $\tq^k_{\ystar}$ is the (discretized) reverse sampling process~(\ref{eq: rev_GP}), and the last inequality holds by the data processing inequality. For the second term, we have
\begin{equation*}
    \TV(\qGpkT, \varepsilon) \lesssim \sqrt{\KL(\qGpk \| \varepsilon)} \exp(-T)
\end{equation*}
Next, we bound the total variation distance between $\qGpk$ and $\tq^k_{\ystar}$, i.e., $\TV(\qGpk, \tq^k_{\ystar})$. Let $Q_{\textup{GP},\ystar,T}^{k,\leftarrow}$ denote the law of the ideal reverse path~(\ref{eq: rev_GP}), which produces $\qGpk$. Define $0=t_0 < t_1 < \cdots < t_J = T$ as the time grid for the discretized backward SDE, where $h=\frac{T}{J} \lesssim L_k^{-1}$ and $t_j = jh$. We denote by $\tilde{Q}_{\textup{GP},\ystar,T}^{k,\leftarrow}$ the path law of the corresponding discretized backward SDE that produces $\tq^k_{\ystar}$, which uses $\tilde{s}_{T-t_{j-1}}(X_{t_{j-1}}^\leftarrow)$ for $t \in [t_{j-1}, t_j]$. Therefore, by the data processing inequality for total variation and Pinsker's inequality, we derive
\begin{equation*}
    \TV(\qGpk, \tq^k_{\ystar}) \le \TV(Q_{\textup{GP},\ystar,T}^{k,\leftarrow}, \tilde{Q}_{\textup{GP},\ystar,T}^{k,\leftarrow}) \le \sqrt{\frac{1}{2} \KL(Q_{\textup{GP},\ystar,T}^{k,\leftarrow} \| \tilde{Q}_{\textup{GP},\ystar,T}^{k,\leftarrow})}.
\end{equation*}
To this end, we first show
\begin{equation}
\begin{aligned}
    & \sum_{j=1}^{J} \E_{Q_{\textup{GP},\ystar,T}^{k,\leftarrow}} \int_{t_{j-1}}^{t_j} \left\| \tilde{s}_{T-t_{j-1}}(X_{t_{j-1}}^\leftarrow) - \nabla \log \qGpktr (X_t^\leftarrow) \right\|^2 \dd t \\
    \lesssim & L_k^2dh + \sqrt{1 + \frac{v_k^2}{\tau^2}} \exp\left( \frac{(1 + r_k)^2}{2\tau^2} \right) L_k^2 h^2 \mathfrak{m}_2^2 \\
    & + \sqrt{1 + \frac{v_k^2}{\tau^2}} \exp\left( \frac{(1 + r_k)^2 + 2}{2\tau^2} \right) \left( \epsilon_\textup{uncond}^2 T + \hat{J}_{\max}^2 \tilde{L}_k^2 \bar \epsilon_{\textup{uncond}}^2 + G_k^2 \bar \epsilon_{\textup{Hess}}^2 \right) \\
    & + J_{\max}^2 \left[ \frac{ (1 + r_k) r_{\nabla, k} }{\tau^2} + \left( \frac{r_k}{\tau^2} + \frac{v_k^2}{\tau^4} \right) B^\star + \frac{1}{2} \left( \frac{ (1 + r_k)^2 }{ \tau^4 } + \frac{ 1 }{\tau^2} \right) v_{\nabla,k} \right]^2 T \\
    & + \frac{(L_\tau r_k + L_{\tau,\var} v_k^2)^2}{c_\tau^2} \left( B_k + \frac{M_k^2}{c_{\tau,k}^2} T \right) + \sqrt{1 + \frac{v_k^2}{\tau^2}} \exp \left( \frac{(1 + r_k)^2 + 1}{2\tau^2} \right) \epsilon_\textup{post-avg}^2.
\end{aligned}
\label{eq: 36}
\end{equation}
Note that
\begin{align}
    & \sum_{j=1}^{J} \E_{Q_{\textup{GP},\ystar,T}^{k,\leftarrow}} \int_{t_{j-1}}^{t_j} \left[ \left\| \tilde{s}_{T-t_{j-1}}(X_{t_{j-1}}^\leftarrow) - \nabla \log \qGpktr (X_t^\leftarrow) \right\|^2 \right] \dd t \nonumber \\
    \lesssim & \sum_{j=1}^{J} \E_{Q_{\textup{GP},\ystar,T}^{k,\leftarrow}} \left[ \int_{t_{j-1}}^{t_j} \left\| \tilde{s}_{T-t_{j-1}}(X_{t_{j-1}}^\leftarrow) - \nabla \log q_{\textup{GP},\ystar,T-t_{j-1}}^k (X_{t_{j-1}}^\leftarrow) \right\|^2 \dd t \right] \label{eq: guidance_err} \\
    & + \sum_{j=1}^{J} \E_{Q_{\textup{GP},\ystar,T}^{k,\leftarrow}}  \left[ \int_{t_{j-1}}^{t_j} \left\| \nabla \log q_{\textup{GP},\ystar,T-t_{j-1}}^k (X_{t_{j-1}}^\leftarrow) - \nabla \log \qGpktr (X_{t_{j-1}}^\leftarrow) \right\|^2 \dd t \right] \label{eq: 2nd_term} \\
    & + \sum_{j=1}^{J} \E_{Q_{\textup{GP},\ystar,T}^{k,\leftarrow}} \left[ \int_{t_{j-1}}^{t_j} \left\| \nabla \log \qGpktr (X_{t_{j-1}}^\leftarrow) - \nabla \log \qGpktr (X_t^\leftarrow) \right\|^2 \dd t \right] \label{eq: 3rd_term}.
\end{align}
\textbf{B.3.2.1. Bounding term~(\ref{eq: 2nd_term})}\quad
For the discretization error, we derive
\begin{align}
    & \E_{Q_{\textup{GP},\ystar,T}^{k,\leftarrow}} \left[ \left\| \nabla \log \frac{q_{\textup{GP},\ystar,T-t_{j-1}}^k}{\qGpktr} (X_{t_{j-1}}^\leftarrow) \right\|^2 \right] \nonumber \\
    \lesssim & \E_{Q_{\textup{GP},\ystar,T}^{k,\leftarrow}} \left[ L_k^2dh + L_k^2 h^2 \| X_{t_{j-1}}^\leftarrow \|^2 + L_k^2 h^2  \| \nabla \log \qGpktr (X_{t_{j-1}}^\leftarrow) \|^2 \right] \nonumber \\
    \lesssim & \E_{Q_{\textup{GP},\ystar,T}^{k,\leftarrow}} \left[ L_k^2dh + L_k^2 h^2 \| X_{t_{j-1}}^\leftarrow \|^2 + L_k^2 h^2  (\| \nabla \log \qGpktr (X_{t}^\leftarrow) \|^2 + L_k^2 \| X_{t_{j-1}}^\leftarrow - X_{t}^\leftarrow \|^2) \right] \nonumber \\
    \lesssim & L_k^2dh + L_k^2 h^2 \sqrt{1 + \frac{v_k^2}{\tau^2}} \exp\left( \frac{(1+r_k)^2}{2\tau^2} \right) \mathfrak{m}_2^2,
\end{align}
where the last inequality holds by $h \lesssim L_k^{-1}$, \citep[Lemmas 10, 11]{chen2023sampling}, and Proposition~\ref{prop: finite_second_moment_gp}. 

\textbf{B.3.2.2. Bounding term~(\ref{eq: 3rd_term})}\quad
We have
\begin{align}
    & \E_{Q_{\textup{GP},\ystar,T}^{k,\leftarrow}} \left[ \left\| \nabla \log \qGpktr (X_{t_{j-1}}^\leftarrow) - \nabla \log \qGpktr (X_t^\leftarrow) \right\|^2 \right] \nonumber \\
    \le & L_k^2 \E_{Q_{\textup{GP},\ystar,T}^{k,\leftarrow}} \left[ \left\| X_{t_{j-1}}^\leftarrow - X_t^\leftarrow \right\|^2 \right] \lesssim L_k^2 (dh + h^2 \sqrt{1 + v_k^2 / \tau^2 } e^{(1+r_k)^2/2\tau^2} \mathfrak{m}_2^2),
\end{align}
where the last inequality holds by ~\citep[Lemma 11]{chen2023sampling} and Proposition~\ref{prop: finite_second_moment_gp}. 

\textbf{B.3.2.3. Bounding term~(\ref{eq: guidance_err})}\quad
For any $t\in [0, T]$, let
\begin{equation*}
    \hat x_0^\theta(x_{t}^\leftarrow, T-t) = \frac{1}{\sqrt{\bar \alpha_{T-t}}}\left( x_{t}^\leftarrow + (1 - \bar \alpha_{T-t}) s_{T-t}^\theta (x_{t}^\leftarrow) \right).
\end{equation*}
We derive
\begin{align*}
    & \sum_{j=1}^{J} \E_{Q_{\textup{GP},\ystar,T}^{k,\leftarrow}} \left[ \int_{t_{j-1}}^{t_j} \left\| \tilde{s}_{T-t_{j-1}}(X_{t_{j-1}}^\leftarrow) - \nabla \log q_{\textup{GP},\ystar,T-t_{j-1}}^k (X_{t_{j-1}}^\leftarrow) \right\|^2 \dd t \right] \\
    \lesssim & \underbrace{\sum_{j=1}^{J} \E_{Q_{\textup{GP},\ystar,T}^{k,\leftarrow}} \left[ \int_{t_{j-1}}^{t_j} \left\| s_{T-t_{j-1}}^\theta(X_{t_{j-1}}^\leftarrow) - \nabla \log q_{T-t_{j-1}} (X_{t_{j-1}}^\leftarrow) \right\|^2 \dd t \right]}_{E_0\textup{ (Unconditional score estimation error)}} \\
    & + \underbrace{\sum_{j=1}^{J} \E_{Q_{\textup{GP},\ystar,T}^{k,\leftarrow}} \left[ \int_{t_{j-1}}^{t_j} \left\| \nabla \log p_{\textup{GP}}^k\left( \ystar \mid \hat x_0^\theta(X_{t_{j-1}}^\leftarrow, T-t_{j-1}) \right) - \nabla \log p_{\textup{GP}}^k\left( \ystar \mid m_{T-t_{j-1}}(X_{t_{j-1}}^\leftarrow) \right) \right\|^2 \dd t \right]}_{E_1\textup{ (Posterior-mean estimation error)}} \\
    & + \underbrace{\sum_{j=1}^{J} \E_{Q_{\textup{GP},\ystar,T}^{k,\leftarrow}} \left[ \int_{t_{j-1}}^{t_j} \left\| \nabla \log p_{\textup{GP}}^k\left( \ystar \mid m_{T-t_{j-1}}(X_{t_{j-1}}^\leftarrow) \right) - \nabla \log \bar p_{\textup{GP},T-t_{j-1}}^k\left( \ystar \mid X_{t_{j-1}}^\leftarrow \right) \right\|^2 \dd t \right]}_{E_2\textup{ (Posterior-averaging error)}}.
\end{align*}
Note that the gradients are taken w.r.t. $X_{t_{j-1}}^\leftarrow$.

\paragraph{Bounding Unconditional Score Estimation Error \texorpdfstring{$E_0$}{E0}}

By Assumption~\ref{as: score_error}, we have
\begin{align*}
    E_0 = & \sum_{j=1}^{J} \E_{Q_{\textup{GP},\ystar,T}^{k,\leftarrow}} \left[ \int_{t_{j-1}}^{t_j} \left\| s_{T-t_{j-1}}^\theta(X_{t_{j-1}}^\leftarrow) - \nabla \log q_{T-t_{j-1}} (X_{t_{j-1}}^\leftarrow) \right\|^2 \dd t \right] \\
    \le & \sqrt{1 + \frac{v_k^2}{\tau^2}} \exp \left( \frac{(1 + r_k)^2 + 1}{2\tau^2} \right) \sum_{j=1}^{J} \E_{Q_{\ystar, T}^\leftarrow} \left[ \int_{t_{j-1}}^{t_j} \left\| s_{T-t_{j-1}}^\theta(X_{t_{j-1}}^\leftarrow) - \nabla \log q_{T-t_{j-1}} (X_{t_{j-1}}^\leftarrow) \right\|^2 \dd t \right] \\
    \le & \sqrt{1 + \frac{v_k^2}{\tau^2}} \exp \left( \frac{(1 + r_k)^2 + 2}{2\tau^2} \right) \epsilon_\textup{uncond}^2 T,
\end{align*}
where the second and the third inequalities hold by Proposition~\ref{prop: dr_gp} and Proposition~\ref{prop: 2}, respectively.

\paragraph{Bounding Posterior-mean Estimation Error \texorpdfstring{$E_1$}{E1}}

Assume
\begin{gather}
    \quad \| J_{\hat x_0^\theta}(x_t, t) \|_\op \le \hat{J}_{\max}, \label{eq: constant_hJ}
\end{gather}
where $\|\cdot\|_\op$ is the operator norm and $J_{\hat x_0^\theta}(x_t, t) := \nabla_{x_t} \hat x_0^\theta(x_t, t)$ is the Jacobian matrix of the posterior-mean prediction map. We have
\begin{align*}
    & \left\| \nabla \log p_{\textup{GP}}^k\left( \ystar \mid \hat x_0^\theta(X_{t_{j-1}}^\leftarrow, T-t_{j-1}) \right) - \nabla \log p_{\textup{GP}}^k\left( \ystar \mid m_{T-t_{j-1}}(X_{t_{j-1}}^\leftarrow) \right) \right\|^2 \\
    \lesssim & \hat{J}_{\max}^2 \tilde{L}_k^2 \| m_{T-t_{j-1}} (X_{t_{j-1}}^\leftarrow) - \hat x_0^\theta(X_{t_{j-1}}^\leftarrow) \|^2 + G_k^2 \| J_{\hat{x}_0^\theta} (X_{t_{j-1}}^\leftarrow, T-t_{j-1}) - J_m (X_{t_{j-1}}^\leftarrow, T-t_{j-1}) \|^2_\op \\
    \lesssim & \frac{\hat{J}_{\max}^2 \tilde{L}_k^2 (1-\bar\alpha_{T-t_{j-1}})^2}{\bar\alpha_{T-t_{j-1}}} \left\| s_{T-{t_{j-1}}}^\theta (X_{t_{j-1}}^\leftarrow) - \nabla \log q_{T-t_{j-1}}(X_{t_{j-1}}^\leftarrow) \right\|^2 \\
    & + \frac{G_k^2 (1-\bar\alpha_{T-t_{j-1}})^2}{\bar\alpha_{T-t_{j-1}}} \left\| \nabla s_{T-{t_{j-1}}}^\theta (X_{t_{j-1}}^\leftarrow) - \nabla^2 \log q_{T-t_{j-1}}(X_{t_{j-1}}^\leftarrow) \right\|^2_\op
\end{align*}
Therefore, we derive
\begin{align*}
    E_1 \lesssim &  \hat{J}_{\max}^2 \tilde{L}_k^2 \sum_{j=1}^{J} \E_{Q_{\textup{GP},\ystar,T}^{k,\leftarrow}} \left[ \frac{(1-\bar\alpha_{T-t_{j-1}})^2}{\bar\alpha_{T-t_{j-1}}} \int_{t_{j-1}}^{t_j} \left\| s_{T-{t_{j-1}}}^\theta (X_{t_{j-1}}^\leftarrow) - \nabla \log q_{T-t_{j-1}}(X_{t_{j-1}}^\leftarrow) \right\|^2 \dd t \right] \\
    & + G_k^2 \sum_{j=1}^{J} \E_{Q_{\textup{GP},\ystar,T}^{k,\leftarrow}} \left[ \frac{(1-\bar\alpha_{T-t_{j-1}})^2}{\bar\alpha_{T-t_{j-1}}} \int_{t_{j-1}}^{t_j} \left\| \nabla s_{T-{t_{j-1}}}^\theta (X_{t_{j-1}}^\leftarrow) - \nabla^2 \log q_{T-t_{j-1}}(X_{t_{j-1}}^\leftarrow) \right\|^2_\op \dd t \right] \\
    \lesssim & \sqrt{1 + \frac{v_k^2}{\tau^2}} \exp \left( \frac{(1 + r_k)^2 + 2}{2\tau^2} \right) \left( \hat{J}_{\max}^2 \tilde{L}_k^2 \bar \epsilon_{\textup{uncond}}^2 + G_k^2 \bar \epsilon_{\textup{Hess}}^2 \right),
\end{align*}
where the last inequality holds by Propositions~\ref{prop: 2} and~\ref{prop: dr_gp}, where we define
\begin{gather}
    \bar \epsilon_{\textup{uncond}}^2 := \sum_{j=1}^{J}  \E_{q_{t_{j}}} \left[ \frac{(1-\bar\alpha_{t_{j}})^2}{\bar\alpha_{t_{j}}} \int_{t_{j-1}}^{t_j} \left\| s_{t_{j}}^\theta (x_{t_{j}}) - \nabla \log q_{t_{j}}(x_{t_{j}}) \right\|^2 \dd t \right] , \label{eq: error_bar_uncond} \\
    \bar \epsilon_{\textup{Hess}}^2 := \sum_{j=1}^{J}  \E_{q_{t_{j}}} \left[ \frac{(1-\bar\alpha_{t_{j}})^2}{\bar\alpha_{t_{j}}} \int_{t_{j-1}}^{t_j} \left\| \nabla s_{t_{j}}^\theta (x_{t_{j}}) - \nabla^2 \log q_{t_{j}}(x_{t_{j}}) \right\|^2_\op \dd t \right] . \label{eq: error_hess}
\end{gather}

\paragraph{Bounding Posterior-averaging Error \texorpdfstring{$E_2$}{E2}}

First, we have
\begin{align*}
    E_2 = & \sum_{j=1}^{J} \E_{Q_{\textup{GP},\ystar,T}^{k,\leftarrow}} \left[ \int_{t_{j-1}}^{t_j} \left\| \nabla \log p_{\textup{GP}}^k\left( \ystar \mid m_{T-t_{j-1}}(X_{t_{j-1}}^\leftarrow) \right) - \nabla \log \bar p_{\textup{GP},T-t_{j-1}}^k\left( \ystar \mid X_{t_{j-1}}^\leftarrow \right) \right\|^2 \dd t \right] \\
    \lesssim & \underbrace{\sum_{j=1}^{J} \E_{Q_{\textup{GP},\ystar,T}^{k,\leftarrow}} \left[ \int_{t_{j-1}}^{t_j} \left\| \nabla \log \frac{\gN\left( \ystar; \mu_k(m_{T-t_{j-1}}(X_{t_{j-1}}^\leftarrow)), \sigma_k^2(m_{T-t_{j-1}}(X_{t_{j-1}}^\leftarrow)) + \tau^2 \right)}{\gN \left( \ystar; \fstar(m_{T-t_{j-1}} (X_{t_{j-1}}^\leftarrow)), \tau^2 \right)} \right\|^2 \dd t \right]}_{E_{2,1}} \\
    & + \underbrace{\sum_{j=1}^{J} \E_{Q_{\textup{GP},\ystar,T}^{k,\leftarrow}} \left[ \int_{t_{j-1}}^{t_j} \left\| \nabla \log \bar p_{T-t_{j-1}} (\ystar|X_{t_{j-1}}^\leftarrow) - \nabla \log \bar p_{\textup{GP},T-t_{j-1}}^k\left( \ystar \mid X_{t_{j-1}}^\leftarrow \right) \right\|^2 \dd t \right]}_{E_{2,2}} \\
    & + \underbrace{\sum_{j=1}^{J} \E_{Q_{\textup{GP},\ystar,T}^{k,\leftarrow}} \left[ \int_{t_{j-1}}^{t_j} \left\| \nabla \log \gN \left( \ystar; \fstar(m_{T-t_{j-1}} (X_{t_{j-1}}^\leftarrow)), \tau^2 \right) - \nabla \log \bar p_{T-t_{j-1}} (\ystar|X_{t_{j-1}}^\leftarrow) \right\|^2 \dd t \right]}_{E_{2,3}}.
\end{align*}
\textbf{- Step A. Bound term $E_{2,1}$.}\quad
Let $J_m(x_t, t) := \nabla_{x_t} m_t(x_t) $ denote the Jacobian matrix of the posterior-mean map. Assume
\begin{gather}
    \| J_m(x_t, t) \|_\op \le J_{\max}, \label{eq: constant_J} \\
    \quad \sup_x \| \nabla \mu_k(x) - \nabla \fstar(x) \| \le r_{\nabla, k} ,  \label{eq: constant_rnk} \\
    \sup_x \| \nabla \sigma_k^2(x) \| \le v_{\nabla, k}. \label{eq: gp-var-grad}
\end{gather}
Let $m = m_{T-t_{j-1}}(X_{t_{j-1}}^\leftarrow)$. By the chain rule, we have
\begin{align*}
    & \nabla \log \frac{\gN\left( \ystar; \mu_k(m_{T-t_{j-1}}(X_{t_{j-1}}^\leftarrow)), \sigma_k^2(m_{T-t_{j-1}}(X_{t_{j-1}}^\leftarrow)) + \tau^2 \right)}{\gN \left( \ystar; \fstar(m_{T-t_{j-1}} (X_{t_{j-1}}^\leftarrow)), \tau^2 \right)} \\
    = & J_m(X_{t_{j-1}}^\leftarrow, T-t_{j-1})^\top \nabla_m \log \frac{ \gN\left( \ystar; \mu_k(m), \sigma_k^2(m) + \tau^2 \right)}{\gN \left( \ystar; \fstar(m), \tau^2 \right)} \\
    = & J_m(X_{t_{j-1}}^\leftarrow, T-t_{j-1})^\top \nabla_m \left( \log \frac{\tau}{\sqrt{\sigma_k^2(m) + \tau^2}} + \frac{( \ystar - \fstar(m) )^2}{2\tau^2} - \frac{( \ystar - \mu_k(m) )^2}{2(\sigma_k^2(m) + \tau^2)} \right) \\
    = & J_m(X_{t_{j-1}}^\leftarrow, T-t_{j-1})^\top  \left( \frac{\ystar - \mu_k}{\sigma_k^2 + \tau^2} \nabla \mu_k - \frac{\ystar - \fstar}{\tau^2} \nabla \fstar + \frac{1}{2} \left( \frac{ ( \ystar - \mu_k )^2 }{ (\sigma_k^2 + \tau^2)^2 } - \frac{ 1 }{ \sigma_k^2 + \tau^2 } \right) \nabla \sigma_k^2 \right).
\end{align*}
Hence, we bound
\begin{align*}
    & \left\| \nabla_m \log \frac{ \gN\left( \ystar; \mu_k(m), \sigma_k^2(m), \tau^2 \right)}{\gN \left( \ystar; \fstar(m), \tau^2 \right)} \right\| \\
    \le & \frac{ (1 + r_k)^2 }{\tau^2} \left\| \nabla \mu_k - \nabla \fstar \right\| + \left( \frac{r_k}{\tau^2} + \frac{v_k^2}{\tau^4} \right) \| \nabla \fstar \| + \frac{1}{2} \left( \frac{ 1 + r_k }{ \tau^4 } + \frac{ 1 }{\tau^2} \right) \| \nabla \sigma_k^2 \| \\
    \le & \frac{ (1 + r_k) r_{\nabla, k} }{\tau^2} + \left( \frac{r_k}{\tau^2} + \frac{v_k^2}{\tau^4} \right) B^\star + \frac{1}{2} \left( \frac{ (1 + r_k)^2 }{ \tau^4 } + \frac{ 1 }{\tau^2} \right) v_{\nabla,k}.
\end{align*}
Therefore, we derive
\begin{equation}
    E_{2,1} \lesssim  J_{\max}^2 \left[ \frac{ (1 + r_k) r_{\nabla, k} }{\tau^2} + \left( \frac{r_k}{\tau^2} + \frac{v_k^2}{\tau^4} \right) B^\star + \frac{1}{2} \left( \frac{ (1 + r_k)^2 }{ \tau^4 } + \frac{ 1 }{\tau^2} \right) v_{\nabla,k} \right]^2 T.
\end{equation}

\textbf{- Step B. Bound term $E_{2,2}$.}\quad
Assume
\begin{equation}
    \sum_{j=1}^{J} \E_{Q_{\textup{GP},\ystar,T}^{k,\leftarrow}} \left\{ \int_{t_{j-1}}^{t_j} \frac{\bar\alpha_{T-t_{j-1}}}{(1 - \bar \alpha_{T-t_{j-1}})^2} \tr\left[\cov_{q_0(\cdot|X_{t_{j-1}}^\leftarrow)}(x_0)]\right] \dd t \right\} \le B_k,
\label{eq: B_k}
\end{equation}
\begin{equation}
    \left\| \nabla \bar p_{\textup{GP},T-t_{j-1}}^k\left( \ystar \mid X_{t_{j-1}}^\leftarrow \right) \right\| \le M_k.
\label{eq: M_k}
\end{equation}
and
\begin{equation}
    p_{\textup{GP},T-t_{j-1}}^k\left( \ystar \mid X_{t_{j-1}}^\leftarrow \right) \ge \frac{1}{\sqrt{2\pi (\tau^2 + v_k^2)}} \exp\left( -\frac{(1 + r_k)^2}{2\tau^2} \right) =: c_{\tau,k}.
\label{eq: constant_ctauk}
\end{equation}
Note that
\begin{gather}
    p_{T-t_{j-1}} (\ystar|X_{t_{j-1}}^\leftarrow) \ge \frac{1}{\sqrt{2\pi} \tau} \exp\left( -\frac{1}{2\tau^2} \right) = c_\tau \\
\end{gather}
We obtain
\begin{align}
    & \left\| \nabla \log \bar p_{T-t_{j-1}} (\ystar|X_{t_{j-1}}^\leftarrow) - \nabla \log \bar p_{\textup{GP},T-t_{j-1}}^k\left( \ystar \mid X_{t_{j-1}}^\leftarrow \right) \right\|^2 \nonumber \\
    = & \left\| \frac{\nabla \bar p_{T-t_{j-1}} (\ystar|X_{t_{j-1}}^\leftarrow)}{\bar p_{T-t_{j-1}} (\ystar|X_{t_{j-1}}^\leftarrow)} - \frac{\nabla \bar p_{\textup{GP},T-t_{j-1}}^k\left( \ystar \mid X_{t_{j-1}}^\leftarrow \right)}{\bar p_{\textup{GP},T-t_{j-1}}^k\left( \ystar \mid X_{t_{j-1}}^\leftarrow \right)} \right\|^2 \nonumber \\
    = & \Bigg\| \frac{\nabla \bar p_{T-t_{j-1}} (\ystar|X_{t_{j-1}}^\leftarrow) - \nabla \bar p_{\textup{GP},T-t_{j-1}}^k\left( \ystar \mid X_{t_{j-1}}^\leftarrow \right)}{\bar p_{T-t_{j-1}} (\ystar|X_{t_{j-1}}^\leftarrow)} \nonumber \\
    & + \nabla \bar p_{\textup{GP},T-t_{j-1}}^k\left( \ystar \mid X_{t_{j-1}}^\leftarrow \right) \left( \frac{1}{\bar p_{T-t_{j-1}} (\ystar|X_{t_{j-1}}^\leftarrow)} - \frac{1}{\bar p_{\textup{GP},T-t_{j-1}}^k\left( \ystar \mid X_{t_{j-1}}^\leftarrow \right)} \right) \Bigg\|^2 \nonumber \\
    \lesssim & \frac{1}{c_\tau^2} \left\| \nabla \bar p_{T-t_{j-1}} (\ystar|X_{t_{j-1}}^\leftarrow) - \nabla \bar p_{\textup{GP},T-t_{j-1}}^k\left( \ystar \mid X_{t_{j-1}}^\leftarrow \right) \right\|^2 \label{eq: e_221} \\
    & + \frac{1}{c_\tau^2 c_{\tau, k}^2} \left\| \nabla \bar p_{\textup{GP},T-t_{j-1}}^k\left( \ystar \mid X_{t_{j-1}}^\leftarrow \right) \right\|^2 \left| \bar p_{T-t_{j-1}} (\ystar|X_{t_{j-1}}^\leftarrow) - \bar p_{\textup{GP},T-t_{j-1}}^k\left( \ystar \mid X_{t_{j-1}}^\leftarrow \right) \right|^2. \label{eq: e_222}
\end{align}

Let 
$$\phi(x) := \gN(\ystar; \fstar(x), \tau^2) - \gN(\ystar; \mu_k(x), \sigma_k^2(x) + \tau^2).$$
\textbf{Sub-step B.1. Bound term~(\ref{eq: e_221}).}\quad
We utilize Lemma~\ref{lem: d_pos_mean} and derive
\begin{align}
    & \nabla \bar p_{T-t_{j-1}} (\ystar|X_{t_{j-1}}^\leftarrow) - \nabla \bar p_{\textup{GP},T-t_{j-1}}^k\left( \ystar \mid X_{t_{j-1}}^\leftarrow \right) \nonumber \\
    = & \nabla \E_{q_0(\cdot | X_{t_{j-1}}^\leftarrow)} \left[ \phi(x_0) \right] \nonumber \\
    = & \cov_{q_0(\cdot | X_{t_{j-1}}^\leftarrow)} \left( \phi(x_0), \nabla \log q_{T-t_{j-1}}(X_{t_{j-1}}^\leftarrow | x_0) \right) \nonumber \\
    = & \cov_{q_0(\cdot | X_{t_{j-1}}^\leftarrow)} \left( \phi(x_0), \frac{\sqrt{\bar\alpha_{T-t_{j-1}}} x_0 - X_{t_{j-1}}^\leftarrow}{1 - \bar \alpha_{T-t_{j-1}}} \right) \nonumber \\
    = & \frac{\sqrt{\bar\alpha_{T-t_{j-1}}}}{1 - \bar \alpha_{T-t_{j-1}}} \cov_{q_0(\cdot | X_{t_{j-1}}^\leftarrow)} \left( \phi(x_0), x_0 \right) \tag{$\cov_{q_0(\cdot|X_{t_{j-1}}^\leftarrow)}(\phi(x_0),X_{t_{j-1}}^\leftarrow)=0$} \\
    = & \frac{\sqrt{\bar\alpha_{T-t_{j-1}}}}{1 - \bar \alpha_{T-t_{j-1}}} \E_{q_0(\cdot|X_{t_{j-1}}^\leftarrow)}\left[ \left( \phi(x_0) - \E_{q_0(\cdot|X_{t_{j-1}}^\leftarrow)}[\phi(x_0)] \right) \left(x_0 - m_{T-t_{j-1}}(X_{t_{j-1}}^\leftarrow) \right) \right] \nonumber
\end{align}
By Cauchy-Schwartz inequality, we further obtain
\begin{align*}
    & \left\| \nabla \bar p_{T-t_{j-1}} (\ystar|X_{t_{j-1}}^\leftarrow) - \nabla \bar p_{\textup{GP},T-t_{j-1}}^k\left( \ystar \mid X_{t_{j-1}}^\leftarrow \right) \right\|^2 \\
    \le & \frac{\bar\alpha_{T-t_{j-1}}}{(1 - \bar \alpha_{T-t_{j-1}})^2} \var_{q_0(\cdot|X_{t_{j-1}}^\leftarrow)} [\phi(x_0)] \tr\left[\cov_{q_0(\cdot|X_{t_{j-1}}^\leftarrow)}(x_0)]\right] \\
    \le & \frac{\bar\alpha_{T-t_{j-1}} }{(1 - \bar \alpha_{T-t_{j-1}})^2} \tr\left[\cov_{q_0(\cdot|X_{t_{j-1}}^\leftarrow)}(x_0)]\right] \E_{q_0(\cdot|X_{t_{j-1}}^\leftarrow)} \left[ \phi^2 (x_0) \right] \\
    \le & \frac{\bar\alpha_{T-t_{j-1}} (L_\tau r_k + L_{\tau, \var} v_k^2 )^2 }{(1 - \bar \alpha_{T-t_{j-1}})^2}  \tr\left[\cov_{q_0(\cdot|X_{t_{j-1}}^\leftarrow)}(x_0)]\right],
\end{align*}
where the last inequality holds by the analysis in Appendix~\ref{sec: b.3.1}, that is,
\begin{align*}
    | \phi(x) | \le & | \gN(\ystar; \fstar(x), \tau^2) - \gN(\ystar; \mu_k(x), \tau^2) | + | \gN(\ystar; \mu_k(x), \tau^2) - \gN(\ystar; \mu_k(x), \sigma_k^2(x) + \tau^2) | \\
    \le & L_\tau r_k + L_{\tau, \var} v_k^2.
\end{align*}
\textbf{Sub-step B.2. Bound term~(\ref{eq: e_222}).}\quad
Note that
\begin{align*}
    & \left| \bar p_{T-t_{j-1}} (\ystar|X_{t_{j-1}}^\leftarrow) - \bar p_{\textup{GP},T-t_{j-1}}^k\left( \ystar \mid X_{t_{j-1}}^\leftarrow \right) \right|^2 \\
    \le &  \E_{q_0(\cdot|X_{t_{j-1}}^\leftarrow)} [\phi^2(x_0)] \\
    \le & ( L_\tau r_k + L_{\tau, \var} v_k^2 )^2.
\end{align*}
Therefore, we derive
\begin{equation}
\begin{aligned}
    E_{2,2} \lesssim & \frac{(L_\tau r_k + L_{\tau,\var} v_k^2)^2}{c_\tau^2} B_k + \frac{(L_\tau r_k + L_{\tau,\var} v_k^2)^2}{c_\tau^2 c_{\tau,k}^2}  M_k^2 T
\end{aligned}
\end{equation}

\textbf{Step C. Bound term $E_{2,3}$.}\quad
Define
\begin{equation}
    \epsilon_{\textup{post-avg}}^2 := \sum_{j=1}^{J} \E_{Q_{\ystar, T}^\leftarrow} \left[ \int_{t_{j-1}}^{t_{j}} \left\| \nabla \log \gN \left( \ystar; \fstar(m_{T-t_{j-1}} (X_{t_{j-1}}^\leftarrow) ), \tau^2 \right) - \nabla \log \bar p_{T-t_{j-1}} (\ystar|X_{t_{j-1}}^\leftarrow) \right\|^2 \dd t \right].
\label{eq: error_post_avg}
\end{equation}
By Proposition~\ref{prop: dr_gp}, we have that
\begin{equation}
    E_{2,3} \le \sqrt{1 + \frac{v_k^2}{\tau^2}} \exp \left( \frac{(1 + r_k)^2 + 1}{2\tau^2} \right) \epsilon_\textup{post-avg}^2
\end{equation}

\subsubsection{Putting Everything Together}

By the same localization and approximation argument as in the proof of~\citep[Theorem 9]{chen2023sampling}, the Girsanov path-space KL bound remains valid without assuming Novikov’s condition. Combining this and Inequality~(\ref{eq: 36}), we have that
\begin{equation*}
    \KL(Q_{\textup{GP},\ystar,T}^{k,\leftarrow} \| \tilde{Q}_{\textup{GP},\ystar,T}^{k,\leftarrow}) \le \sum_{j=1}^{J} \E_{Q_{\textup{GP},\ystar,T}^{k,\leftarrow}} \int_{t_{j-1}}^{t_j} \left\| \tilde{s}_{T-t_{j-1}}(X_{t_{j-1}}^\leftarrow) - \nabla \log \qGpktr (X_t^\leftarrow) \right\|^2 \dd t.
\end{equation*}
Therefore, we derive
\begin{align*}
    \TV(q_\ystar, \hq^k_{\ystar}) \le & \TV(q_\ystar, \qGpk) + \TV(\qGpk, \tq^k_{\ystar}) + \TV(\qGpkT, \varepsilon) \\
    \le & \TV(q_\ystar, \qGpk) + \sqrt{\KL(\qGpk \| \varepsilon)} \exp(-T) + \sqrt{ \KL(Q_{\textup{GP},\ystar,T}^{k,\leftarrow} \| \tilde{Q}_{\textup{GP},\ystar,T}^{k,\leftarrow}) } \\
    \lesssim & \frac{L_\tau r_k + L_{\tau,\var} v_k^2 }{c_\tau} + \sqrt{\KL(\qGpk \| \varepsilon)} \exp(-T)\\
    & + \sqrt{ J_{\max}^2 \left[ \frac{ (1 + r_k) r_{\nabla, k} }{\tau^2} + \left( \frac{r_k}{\tau^2} + \frac{v_k^2}{\tau^4} \right) B^\star + \frac{1}{2} \left( \frac{ (1 + r_k)^2 }{ \tau^4 } + \frac{ 1 }{\tau^2} \right) v_{\nabla,k} \right]^2 T } \\
    & + \frac{L_\tau r_k + L_{\tau,\var} v_k^2}{c_\tau} \sqrt{B_k + \frac{M_k^2}{c_{\tau,k}^2} T} \\
    & + \left( 1 + \frac{v_k^2}{\tau^2} \right)^{1/4} \exp\left( \frac{(1 + r_k)^2 + 2}{4\tau^2} \right) \sqrt{ \epsilon_\textup{uncond}^2 T + \hat{J}_{\max}^2 \tilde{L}_k^2 \bar \epsilon_{\textup{uncond}}^2 + G_k^2 \bar \epsilon_{\textup{Hess}}^2 } \\
    & +  \left( 1 + \frac{v_k^2}{\tau^2} \right)^{1/4} \exp\left( \frac{(1 + r_k)^2 + 1}{4\tau^2} \right) \epsilon_\textup{post-avg} \\
    & + \sqrt{ L_k^2dh + \sqrt{1 + \frac{v_k^2}{\tau^2}} \exp\left( \frac{(1 + r_k)^2}{2\tau^2} \right) L_k^2 h^2 \mathfrak{m}_2^2 }
\end{align*}

\subsection{Auxiliary Proofs}

\begin{proposition}[Density ration between target and data distribution]
Under Assumption~\ref{as: bounded_reward} and the target reward $\ystar \in [0,1]$, we have that
\begin{equation}
    \left\| \frac{\dd q_\ystar}{\dd \qd} \right\|_\infty, \left\| \frac{\dd \qd}{\dd q_\ystar} \right\|_\infty \le \exp\left( \frac{1}{2\tau^2} \right).
\end{equation}
\begin{proof}
By definition, we have
\begin{align*}
    \frac{q_\ystar (x)}{\qd (x)} = \frac{\gN(\ystar; \fstar (x), \tau^2)}{Z_\ystar}.
\end{align*}
Since $\ystar, \fstar(x) \in [0,1]$, we have
\begin{gather*}
    \frac{1}{\sqrt{2\pi} \tau}\exp\left(-\frac{1}{2\tau^2}\right) \le \gN(\ystar; \fstar (x), \tau^2) \le \frac{1}{\sqrt{2\pi} \tau},\\
    \frac{1}{\sqrt{2\pi} \tau}\exp\left(-\frac{1}{2\tau^2}\right) \le Z_\ystar \le \frac{1}{\sqrt{2\pi} \tau},
\end{gather*}
which concludes the proof.
\end{proof}
\label{prop: 1}
\end{proposition}

\begin{proposition}
Under Assumptions~\ref{as: score_error}, we have
\begin{equation*}
    \E_{x_t \sim q_{\ystar,t}} \left[ \left\| s_t^\theta(x_t) - \nabla \log q_t (x_t) \right\|^2 \right] \le \exp\left( \frac{1}{2\tau^2} \right) \epsUncond^2
\end{equation*}
\label{prop: 2}
\end{proposition}

\begin{proposition}[Density ratio between target and GP-conditioned distribution]
\label{prop: dr_gp}
For any $t \in [0, T]$, we have that
\begin{equation}
    \left\| \frac{\qGpkt}{q_{\ystar, t}} \right\|_\infty \le \sqrt{1 + \frac{v_k^2}{\tau^2}} \exp \left( \frac{(1 + r_k)^2 + 1}{2\tau^2} \right).
\label{eq: c_k,ystar}
\end{equation}
\begin{proof}
We first prove the density-ratio bound at time $t=0$, and then use the fact that the forward noising kernel preserves the upper bounds.

\textbf{Case 1. $t=0$}.\quad
First, we have that
\begin{equation*}
    \frac{\qGpk(x)}{q_{\ystar}(x)} = \frac{Z_\ystar}{Z_k}\frac{\gN(\ystar; \mu_k(x), \sigma_k^2(x) + \tau^2)}{\gN(\ystar; \fstar(x), \tau^2)}.
\end{equation*}
Note that
\begin{align*}
    \frac{\gN(\ystar; \mu_k(x), \sigma_k^2(x) + \tau^2)}{\gN(\ystar; \fstar(x), \tau^2)} = \frac{\tau}{\sqrt{\tau^2 + \sigma_k^2(x)}} \exp \left( \frac{( \ystar - \fstar(x) ) ^2 }{ 2 \tau^2 }  - \frac{ (\ystar - \mu_k(x))^2 }{ 2 (\tau^2 + \sigma_k^2 (x) ) }\right) \le \exp \left( \frac{1}{2\tau^2} \right).
\end{align*}
On the other hand, we obtain
\begin{align*}
    \frac{\gN(\ystar; \fstar(x), \tau^2)}{\gN(\ystar; \mu_k(x), \sigma_k^2(x) + \tau^2)} = & \frac{\sqrt{\tau^2 + \sigma_k^2(x)}}{\tau} \exp \left( \frac{ (\ystar - \mu_k(x))^2 }{ 2 (\tau^2 + \sigma_k^2 (x) ) } - \frac{( \ystar - \fstar(x) ) ^2 }{ 2 \tau^2 } \right) \\
    \le & \sqrt{1 + \frac{v_k^2}{\tau^2}} \exp \left( \frac{(1 + r_k)^2}{2\tau^2} \right).
\end{align*}
Hence, we derive
\begin{align*}
    Z_\ystar = & \int \qd(x) \gN(\ystar; \fstar(x), \tau^2) \dd x \\
    = & \int \qd(x) \gN(\ystar; \mu_k(x), \sigma_k^2(x) + \tau^2) \frac{\gN(\ystar; \fstar(x), \tau^2)}{\gN(\ystar; \mu_k(x), \sigma_k^2(x) + \tau^2)} \dd x \\
    \le & \sqrt{1 + \frac{v_k^2}{\tau^2}} \exp \left( \frac{(1 + r_k)^2}{2\tau^2} \right) \int \qd(x) \gN(\ystar; \mu_k(x), \sigma_k^2(x) + \tau^2) \dd x = \sqrt{1 + \frac{v_k^2}{\tau^2}} \exp \left( \frac{(1 + r_k)^2}{2\tau^2} \right) Z_k.
\end{align*}
Therefore, we prove
\begin{equation*}
    \frac{\qGpk(x)}{q_{\ystar}(x)} \le \sqrt{1 + \frac{v_k^2}{\tau^2}} \exp \left( \frac{(1 + r_k)^2 + 1}{2\tau^2} \right).
\end{equation*}

\textbf{Case 2. $t \in [0, T]$.}\quad
Note that
\begin{align}
    \qGpkt(x_t) = & \int \qGpkt(x_t|x_0) \qGpk(x_0) \dd x_0 \nonumber \\
    = & \int q_{\ystar, t}(x_t|x_0) q_\ystar(x_0) \frac{\qGpk(x_0)}{q_\ystar(x_0)} \dd x_0 \tag{$q_{\ystar, t}(x_t|x_0) = \qGpkt(x_t|x_0)$} \\
    \le & \sqrt{1 + \frac{v_k^2}{\tau^2}} \exp \left( \frac{(1 + r_k)^2+1}{2\tau^2} \right) \int q_{\ystar, t}(x_t|x_0) q_\ystar(x_0) \dd x_0 \nonumber \\
    = & \sqrt{1 + \frac{v_k^2}{\tau^2}} \exp \left( \frac{(1 + r_k)^2+1}{2\tau^2} \right) q_{\ystar, t}(x_t), \nonumber
\end{align}
which concludes the proof.
\end{proof}
\end{proposition}

\begin{proposition}
Under Assumptions~\ref{as: bounded_reward} and~\ref{as: second_moment}, $\qGpk$ has a finite second moment.
\begin{proof}
Recall that
\begin{equation}
    \qGpk(x) = \frac{\qd(x) \gN(\ystar; \mu_k(x), \sigma_k^2(x) + \tau^2)}{Z_k},
\end{equation}
where $Z_k = \int \qd(x) \gN(\ystar; \mu_k(x), \sigma_k^2(x) + \tau^2) \dd x$ is the normalizer. Then, we have
\begin{align*}
    \E_{\qGpk}\left[ \|x\|^2 \right] = \frac{1}{Z_k} \int \| x \|^2 \qd(x) \gN(\ystar; \mu_k(x), \sigma_k^2(x) + \tau^2) \dd x \le \frac{1}{Z_k \sqrt{2\pi} \tau} \E_\qd \left[ \|x\|^2 \right].
\end{align*}
Further, since $|\mu_k(x) - \fstar (x)| \le r_k$ and $ \sigma_k^2 \le v_k^2 $, we have that $Z_k \ge \frac{1}{\sqrt{2\pi (\tau^2 + v_k^2 )}} \exp(-(1+r_k)^2/2\tau^2)$. Therefore, we derive
\begin{equation*}
    \E_{\qGpk}\left[ \|x\|^2 \right] \le \sqrt{1 + \frac{v_k^2}{\tau^2}} \exp\left( \frac{(1 + r_k)^2}{2\tau^2} \right)\E_\qd \left[ \|x\|^2 \right] \le \sqrt{1 + \frac{v_k^2}{\tau^2}} \exp\left( \frac{(1 + r_k)^2}{2\tau^2} \right) \mathfrak{m}_2^2 ,
\end{equation*}
which concludes the proof.
\end{proof}
\label{prop: finite_second_moment_gp}
\end{proposition}

\begin{lemma}[Derivative of posterior mean]
\label{lem: d_pos_mean}
Let $\phi: \gX \to \sR$ be a function with bounded values. For $t>0$, we have that
\begin{equation}
    \nabla_{x_t} \E_{q_0(\cdot|x_t)} \left[ \phi(x_0) | x_t \right] = \cov_{q_0(\cdot|x_t)} \left( \phi(x_0), \nabla_{x_t} \log q_t(x_t|x_0) \right).
\end{equation}
\begin{proof}
Recall that
\begin{align*}
    q_0(x_0|x_t) = \frac{\qd(x_0) q_t(x_t|x_0)}{q_t(x_t)},
\end{align*}
where $q_t(x_t|x_0) = \gN(x_t; \sqrt{\bar \alpha_t} x_0, (1 - \bar\alpha_t) I)$. Define $F(x_t) := \E_{q_0(\cdot|x_t)} [\phi(x_0) | x_t] = \int \phi(x_0) q_0(x_0|x_t) \dd x_0$. We have that
\begin{align*}
    \nabla F(x_t) = \frac{\nabla A(x_t)}{q_t(x_t)} - \frac{A(x_t)}{q_t(x_t)} \nabla \log q_t(x_t),
\end{align*}
where $A(x_t) := \int \phi(x_0) \qd(x_0) q_t(x_t|x_0) \dd x_0$. Since $\phi$ is bounded and the Gaussian kernel $q_t(x_t|x_0)$ is smooth in $x_t$, we can interchange the derivative and the integral and obtain that
\begin{gather*}
    \nabla A(x_t) = \int \phi(x_0) \qd(x_0) \nabla q_t(x_t|x_0) \dd x_0 = \int \phi(x_0) \qd(x_0) q_t(x_t|x_0) \nabla \log q_t(x_t|x_0) \dd x_0.
\end{gather*}
Hence, we have that
\begin{align*}
    \frac{\nabla A(x_t)}{q_t(x_t)} = \int \phi(x_0) \nabla \log q_t(x_t|x_0) q_0(x_0|x_t) \dd x_0 = \E_{q_0(\cdot|x_t)} [\phi(x_0) \nabla \log q_t(x_t|x_0)].
\end{align*}
On the other hand, we have that
\begin{align*}
    \frac{A(x_t)}{q_t(x_t)} \nabla \log q_t(x_t) = & \nabla \log q_t(x_t) \int \phi(x_0) q_0(x_0|x_t) \dd x_0 = \nabla \log q_t(x_t) \E_{q_0(\cdot|x_t)} [\phi(x_0)].
\end{align*}
Moreover, note that
\begin{align*}
    \nabla \log q_t(x_t) = \frac{\nabla q_t(x_t)}{q_t(x_t)} = \frac{\int \qd(x_0) \nabla q_t(x_t|x_0) \dd x_0}{q_t(x_t)} = \E_{q_0(\cdot|x_t)} [\nabla \log q_t(x_t|x_0)],
\end{align*}
where we utilize the fact that $\nabla_{x_t} q_t(x_t|x_0) = q_t(x_t|x_0) \nabla_{x_t} \log q_t(x_t|x_0)$ to derive the last equation. We conclude the proof.
\end{proof}
\end{lemma}

\section{Detailed Implementation}

\textbf{{\subA} \& \subB}\quad At each round, \textsc{BReaD} fits a GP/DKL surrogate using the observed labeled samples collected so far. Given the current noisy sample at a denoising step, it first predicts the posterior mean~(\ref{eq: predicted_mean}) and extracts a differentiable feature representation from the predicted clean sample. The GP posterior is then evaluated on this representation to further compute target-conditioned likelihood in {\subA} or the upper confidence bound (UCB) in {\subB}. The resulting guidance gradient is differentiated with respect to the noisy sample by back-propagating through the feature extractor and the diffusion model.

\textbf{\mlp}\quad
The MLP-guided baseline uses a lightweight multilayer perceptron (MLP) to predict the black-box objective value from the same feature representation used by \textsc{BReaD}. The MLP shares the same architecture as the feature map within our deep kernel learning (DKL) surrogate, but directly outputs a scalar reward prediction. At each denoising step, it first predicts the posterior mean~(\ref{eq: predicted_mean}) of the current noisy sample $X_t^\leftarrow$. This estimate is then forwarded to extract the feature input to the MLP surrogate. Then, the guidance gradient is computed via backpropagation and used to modify the noise estimate in the forward universal guidance style.

\textbf{\gpucb}\quad
The GP-UCB baseline follows a generate-and-select strategy. At each iteration, it first generates a batch of \num{10} candidate samples from the pre-trained diffusion model. The UCB values of these samples are computed from a GP surrogate that is fitted on the observed data so far. The candidate with the highest UCB value is selected for evaluation at the current iteration.

\textbf{\fwd}\quad
The implementation of the forward universal guidance baseline follows\footnote{\url{https://github.com/arpitbansal297/Universal-Guided-Diffusion}}~\citep{bansal2024universal}. At every timestep, the current noisy sample is used to predict the posterior mean~\ref{eq: predicted_mean}. This estimate is passed to the true objective function to compute the exact reward gradient for guidance.

\textbf{Input features for {\subA}, {\subB}, {\mlp}, and \gpucb.}\quad
For a given experiment, all four methods use the same feature representation to ensure a consistent comparison. For image generation tasks in Section~\ref{sec: 4.3}, the feature is the \num{768}-dimensional CLIP embedding of the predicted clean image. For the ligand optimization task in Section~\ref{sec: ligand}, we obtain the \num{256}-dimensional feature by first extracting the hidden representation immediately before the EGNN output projection and then mean-pooling the representations over the ligand atoms. For the WPQH experiment in Section~\ref{sec: wpqh exp}, we parameterize the generated sequence of six actuators into \num{21} parameters and use these parameters as the surrogate input.

\section{Additional Details in Section~\ref{sec: wpqh exp}: Results for Controllable Nuclear Fusion}
\label{sec: C}

\paragraph{Controllable nuclear fusion and Wide-Pedestal QH.}

Nuclear fusion is a promising path toward sustainable energy, and tokamaks are among the leading candidates for producing controlled thermonuclear fusion power. Conventional approaches to plasma control typically combine precomputed actuator trajectories, such as coil-current programs, with feedback controllers designed for specific target quantities. These control systems have demonstrated reliable performance within a wide range of discharges, but their design requires substantial engineering effort and expertise, especially for advanced plasma scenarios. One such example is accessing the Wide-Pedestal QH~(WPQH) mode. The WPQH mode is an advanced plasma-confinement regime characterized by a wide edge transport barrier that enables high energy confinement while avoiding damaging edge-localized modes (ELMs).

\paragraph{Modeling DIII-D Tokamak dynamics using an RPNN ensemble.}

We follow~\citep{char2024shotpredictionsdiiidtokamak} and model plasma dynamics in the DIII-D Tokamak using an ensemble of recurrent probabilistic neural networks (RPNNs). Each RPNN autoregressively predicts the distribution of the change in plasma state between consecutive timesteps, conditioned on the current plasma state, actuator inputs, and a recurrent hidden state that summarizes the preceding trajectory. The predicted state difference is then added to the current state to obtain the next-state prediction. The network architecture is shown below. We train an ensemble of \num{25} RPNNs to predict \num{12} plasma states, consisting of \num{10} scalar states and two plasma profiles, with each profile represented by four PCA components, from a sequence of \num{14} actuators parameterized by 19 input variables. To evaluate the predictive accuracy of the learned dynamics model, we report the explained variance (EV) of the predicted timestep differences, i.e., the change from the current state to the next state, for each state variable in Table~\ref{tab: rpnn_ev}.

\noindent\rule{\linewidth}{0.08em}
\vspace{0.2em}
\textbf{Network Architecture:}
\begin{itemize}
    \item \textbf{Encoder:}
    \begin{itemize}
        \item Fully Connected (FC) layer: \texttt{input\_dim} $\times$ 512
        \item FC layer: 512 $\times$ 512
    \end{itemize}
    
    \item \textbf{Memory Unit:}
    \begin{itemize}
        \item Gated Recurrent Unit (GRU) block: 512 $\times$ 256
    \end{itemize}

    \item \textbf{Decoder (with residual connections between FC layers):}
    \begin{itemize}
        \item FC layer: 256 $\times$ 512
        \item FC layers: 512 $\times$ 512 (repeated 8 times)
        \item FC layer: 512 $\times$ 128
    \end{itemize}

    \item \textbf{Output Heads:}
    \begin{itemize}
        \item Mean head: 128 $\times$ \texttt{output\_dim}
        \item Log-variance head: 128 $\times$ \texttt{output\_dim}
    \end{itemize}
\end{itemize}
\vspace{0.2em}
\noindent\rule{\linewidth}{0.08em}

\begin{table}[!ht]
    \centering
    \caption{Per-variable explained variance (EV) of the RPNN ensemble trained on historical DIII-D experimental data. Higher values indicate better agreement with held-out real trajectories.}
    \label{tab:wpqh-dynamics}
    \small
    \setlength{\tabcolsep}{6pt}
    \begin{tabular}{l c l c}
        \toprule
        Variable & Ensemble EV & Variable & Ensemble EV \\
        \midrule
        $\beta_N$              & 0.7983 & denv1       & 0.4912 \\
        $\beta_p$              & 0.8137 & temp\_pca1  & 0.6892 \\
        $W_{\mathrm{MHD}}$     & 0.8012 & temp\_pca2  & 0.5145 \\
        prmtan\_tewid          & 0.3132 & temp\_pca3  & 0.3936 \\
        prmtan\_newid          & 0.2460 & temp\_pca4  & 0.4799 \\
        prmtan\_teped          & 0.3803 & dens\_pca1  & 0.6918 \\
        prmtan\_neped          & 0.4126 & dens\_pca2  & 0.5333 \\
        FS04\_max              & 0.5999 & dens\_pca3  & 0.5380 \\
        denr0                  & 0.4538 & dens\_pca4  & 0.5692 \\
        \midrule
        \multicolumn{3}{l}{\textbf{Average Ensemble EV}} 
        & \textbf{0.5400} \\
        \bottomrule
    \end{tabular}
\label{tab: rpnn_ev}
\end{table}
\vspace{-1em}

\begin{table}[!b]
\centering
\setlength{\tabcolsep}{6pt}
\renewcommand{\arraystretch}{1.25}
\begin{tabular}{l >{\raggedright\arraybackslash}p{0.36\linewidth} ccc}
\toprule
\textbf{Signal Group} & \textbf{Signals} &
\textbf{Actuator} & \textbf{Controllable} & \textbf{State} \\
\midrule

\multirow{10}{*}{\textbf{Scalar States}}
  & $\beta_N$ (Normalized Plasma Pressure)
  & \xmark & \xmark & \cmark \\
  & $\beta_p$ (Poloidal Beta)
  & \xmark & \xmark & \cmark \\
  & MHD Stored Energy ($W_{\mathrm{MHD}}$)
  & \xmark & \xmark & \cmark \\
  & Electron Temperature Pedestal Width (prmtan\_tewid)
  & \xmark & \xmark & \cmark \\
  & Electron Density Pedestal Width (prmtan\_newid)
  & \xmark & \xmark & \cmark \\
  & Electron Temperature Pedestal Height (prmtan\_teped)
  & \xmark & \xmark & \cmark \\
  & Electron Density Pedestal Height (prmtan\_neped)
  & \xmark & \xmark & \cmark \\
  & FilterScope ($\mathrm{FS04}_{\max}$)
  & \xmark & \xmark & \cmark \\
  & \texttt{denr0}
  & \xmark & \xmark & \cmark \\
  & \texttt{denv1}
  & \xmark & \xmark & \cmark \\
\midrule

\multirow{2}{*}{\textbf{Profile States}}
  & Electron Temperature Profile ($T_e$)
  & \xmark & \xmark & \cmark \\
  & Electron Density Profile ($n_e$)
  & \xmark & \xmark & \cmark \\
\midrule

\multirow{4}{*}{\textbf{Plasma Control}}
  & NBI Total Power ($P_{\mathrm{inj}}$)
  & \cmark & \cmark & \xmark \\
  & NBI Torque ($T_{\mathrm{inj}}$)
  & \cmark & \cmark & \xmark \\
  & Plasma Current ($I_p$)
  & \cmark & \cmark & \xmark \\
  & Safety Factor ($q_{95}$)
  & \cmark & \cmark & \xmark \\
\midrule

\multirow{6}{*}{\textbf{Shape Variables}}
  & Separatrix Distance ($dR_{\mathrm{sep}}$)
  & \cmark & \cmark & \xmark \\
  & Bottom Triangularity (tribot)
  & \cmark & \xmark & \xmark \\
  & Top Triangularity (tritop)
  & \cmark & \xmark & \xmark \\
  & Elongation ($\kappa$)
  & \cmark & \xmark & \xmark \\
  & Minor Radius ($a_{\mathrm{minor}}$)
  & \cmark & \xmark & \xmark \\
  & Radial Positions of Magnetic Axis ($R_\mathrm{axis}$)
  & \cmark & \xmark & \xmark \\
  & Vertical Positions of Magnetic Axis ($Z_\mathrm{axis}$)
  & \cmark & \xmark & \xmark \\
\midrule

\multirow{6}{*}{\makecell[l]{\textbf{Electron Cyclotron}\\\textbf{Heating (ECH)}}}
  & \texttt{ec\_j\_eccd\_param1}
  & \cmark & \cmark & \xmark \\
  & \texttt{ec\_j\_eccd\_param2}
  & \cmark & \cmark & \xmark \\
  & \texttt{ec\_j\_eccd\_param3}
  & \cmark & \cmark & \xmark \\
  & \texttt{ec\_qrfe\_param1}
  & \cmark & \cmark & \xmark \\
  & \texttt{ec\_qrfe\_param2}
  & \cmark & \cmark & \xmark \\
  & \texttt{ec\_qrfe\_param3}
  & \cmark & \cmark & \xmark \\
\midrule

\multirow{2}{*}{\textbf{Other Actuators}}
  & Toroidal Field ($B_T$)
  & \cmark & \xmark & \xmark \\
  & Gas Puffing (\texttt{gasA})
  & \cmark & \xmark & \xmark \\
\bottomrule
\end{tabular}

\caption{Plasma state and actuator signals used for modeling the DIII-D
tokamak dynamics in the WPQH experiments. The controllable column indicates
actuator signals that can be optimized when searching for WPQH access strategies.}
\label{tab: rpnn}
\end{table}

\paragraph{Training WPQH mode detector.} 

To evaluate whether a plasma trajectory accesses the WPQH mode, we train a binary classifier to identify WPQH operation at each timestep. Our labeled dataset contains \num{477} DIII-D shots that successfully accessed WPQH and \num{308} shots that did not. After data preprocessing, including removing shots with missing required signals, \num{434} shots remain for training the detector. The WPQH detector is a multilayer perceptron (MLP) that predicts whether the plasma is in the WPQH mode at a given timestep using a local temporal window of plasma states and actuator signals spanning \num{60} ms before and after the target timestep. At each timestep, we use five scalar plasma states ($\beta_N$, $W_{\mathrm{MHD}}$, $\beta_p$, electron-temperature pedestal width (prmtan\_tewid), and electron-density pedestal width (prmtan\_newid)), the first two PCA components of the electron-temperature ($T_e$) and electron-density ($n_e$) profiles, and five actuator signals (NBI power $P_\mathrm{inj}$, plasma current ($I_p$), NBI torque ($T_\mathrm{inj}$), $q_{95}$, and separatrix distance ($dR_\mathrm{sep}$)). The resulting window is flattened and passed through two fully connected hidden layers with \num{128} and \num{64} units and ReLU activations, followed by a binary classification head. 

We evaluate the detector under two settings. Real-data detection directly applies the detector to measured plasma trajectories and achieves \num{85.49}\% accuracy and \num{83.98}\% precision on WPQH predictions. Rollout-data detection first replays the experimental shots using the learned RPNN dynamics model and then applies the same detector to the simulated trajectories. Despite the additional error introduced by the rollouts, the detector achieves \num{72.46}\% accuracy and \num{69.21}\% precision on simulated trajectories. These results indicate that the detector retains reasonable predictive performance on model-generated rollouts, making it reasonable to use it to estimate the WPQH ratio in our simulation experiments.

\paragraph{Pre-training diffusion model on DIII-D dataset.}

We pre-train a conditional diffusion model~(DM) on historical DIII-D trajectories to generate future plasma states and actuator sequences. Each training sample consists of a trajectory of \num{200} timesteps~(\num{4000}ms), where the first \num{10} timesteps~(\num{200}ms) are provided as a clean conditioning prefix, and the model generates the remaining \num{190} timesteps~(\num{3800}ms). The DM jointly models plasma states and actuator sequences~(in Table~\ref{tab: rpnn}), and it enables both joint trajectory generation and conditional generation, i.e., predicting the corresponding plasma dynamics given the actuator trajectories. We parameterize the diffusion model using a temporal Diffusion Transformer~\citep[DiT]{esser2024scaling}. At each timestep, the plasma state and actuator signals are represented by separate tokens with dedicated input projections and learnable type embeddings. The model consists of eight transformer blocks with a hidden dimension of \num{256} and eight attention heads. Each block contains masked self-attention across the generated trajectory, cross-attention to the clean conditioning prefix, and a feed-forward network. The diffusion timestep is incorporated through adaptive layer normalization~(AdaLN). For forward prediction, actuator signals are kept fixed while the DM denoises the plasma-state sequence. We train the model using an exponential noise schedule with \num{1}k diffusion steps and an OU forward process. To evaluate the learned dynamics, we condition the model on the initial plasma state and actuator trajectory and measure the explained variance (EV) between the generated and real plasma trajectories. 

\section{Additional Numerical Results}
\label{sec: additional results}

\begin{figure}[!h]
    \centering
    \begin{subfigure}[t]{0.25\textwidth}
        \centering
        \includegraphics[height=3.4cm]{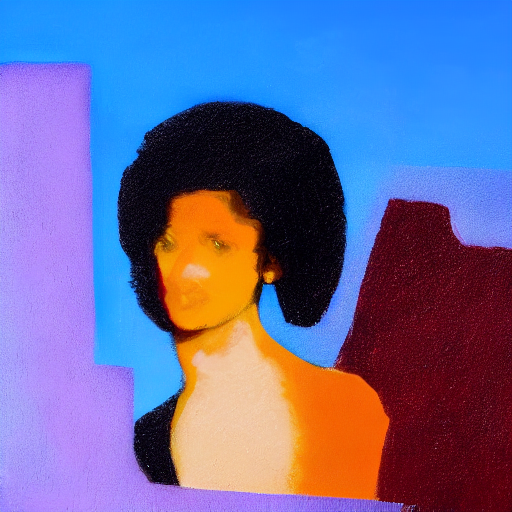}
        \caption{{\subA}}
    \end{subfigure}
    \hfill
    \begin{subfigure}[t]{0.25\textwidth}
        \centering
        \includegraphics[height=3.4cm]{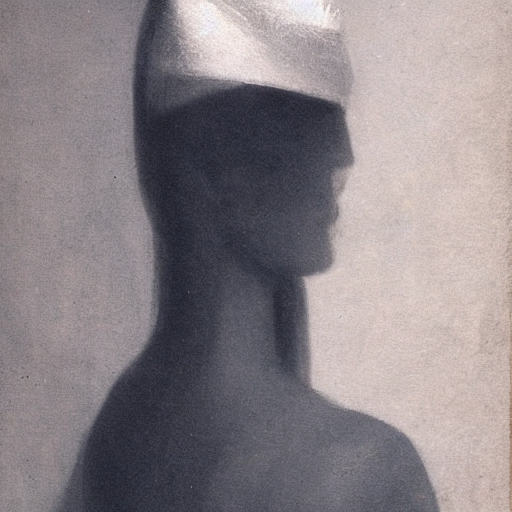}
        \caption{{\subB}}
    \end{subfigure}
    \hfill
    \begin{subfigure}[t]{0.43\textwidth}
        \centering
        \includegraphics[height=4.5cm]{figs/add_se/losses_avg_portrait_cubist_se.pdf}
        \caption{Best-so-far style loss}
    \end{subfigure}

    \caption{Style guidance on Stable Diffusion v1.5 to match the text prompt, \textit{A portrait of a woman}, and the style of Pablo Picasso's painting, \textit{Brick Factory at Tortosa}. Results are averaged over 10 trials.}
    \label{fig:style guidance 3}
\end{figure}

\begin{figure}[!h]
    \centering    
    \begin{subfigure}[b]{0.25\textwidth}
        \centering
        \includegraphics[height=3.4cm]{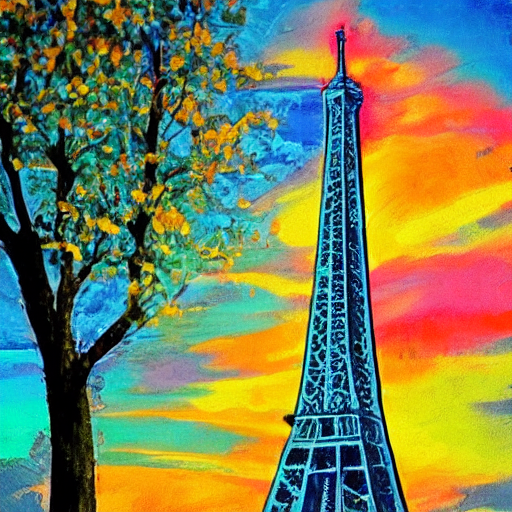}
        \caption{{\subA}}
        \label{fig:style guidance 1(b)}
    \end{subfigure}
    \hfill
    \begin{subfigure}[b]{0.25\textwidth}
        \centering
        \includegraphics[height=3.4cm]{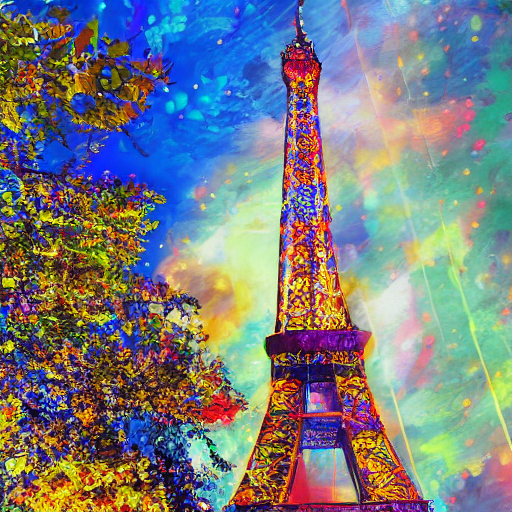}
        \caption{{\subB}}
        \label{fig:style guidance 1(a)}
    \end{subfigure}
    \hfill
    \begin{subfigure}[b]{0.43\textwidth}
        \centering
        \includegraphics[height=4.5cm]{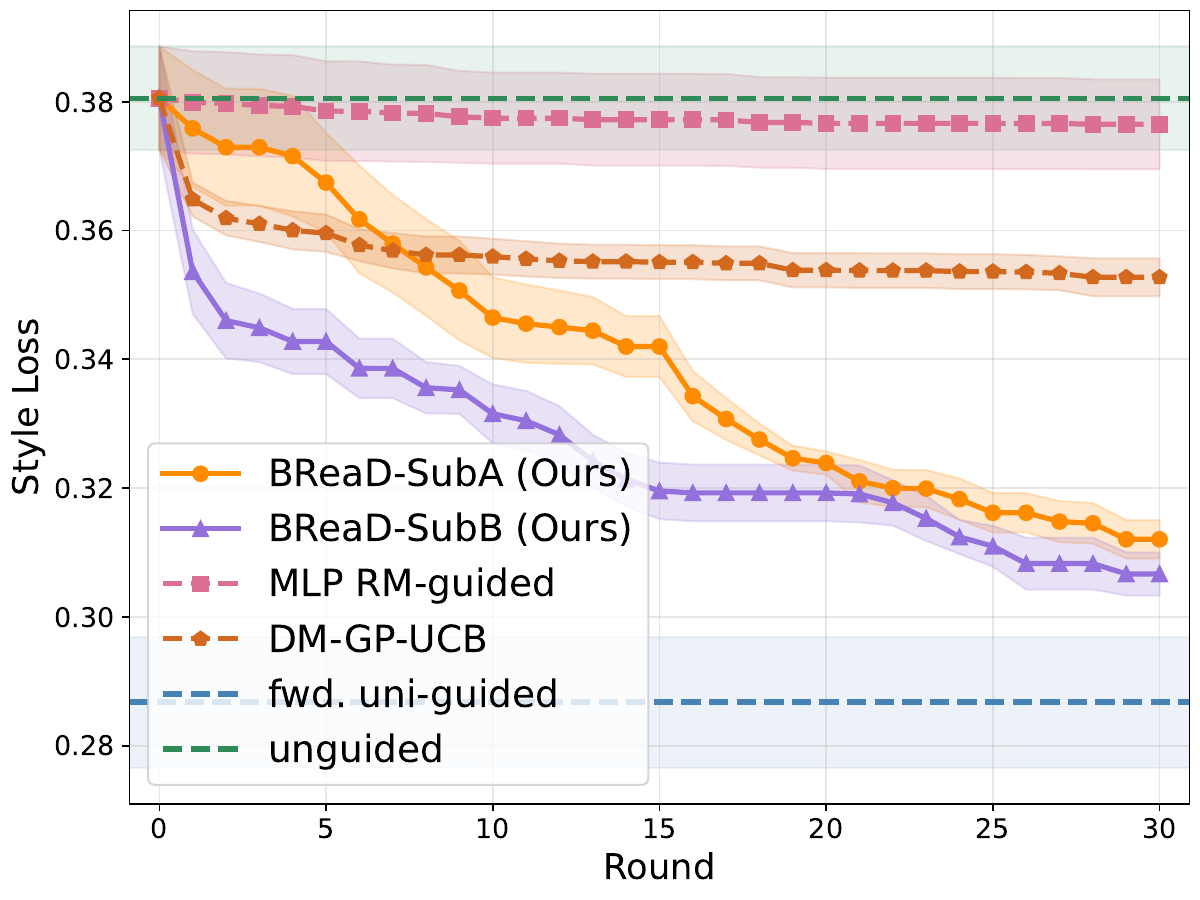}
        \caption{Best-so-far style loss}
        \label{fig:style guidance 1(c)}
    \end{subfigure}

    \caption{Style guidance on Stable Diffusion v1.5 to match the text prompt, \textit{A colorful photo of an Eiffel Tower}, and the style of Vincent van Gogh's painting, \textit{The Starry Night}. Results are averaged over 10 trials.}
    \label{fig:style guidance 1}
\end{figure}

\begin{figure}[ht]
    \centering
    \begin{subfigure}[t]{0.25\textwidth}
        \centering
        \includegraphics[height=3.4cm]{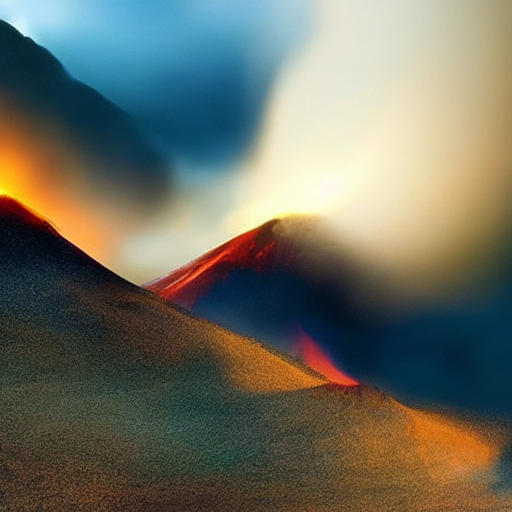}
        \caption{{\subA}}
    \end{subfigure}
    \hfill
    \begin{subfigure}[t]{0.25\textwidth}
        \centering
        \includegraphics[height=3.4cm]{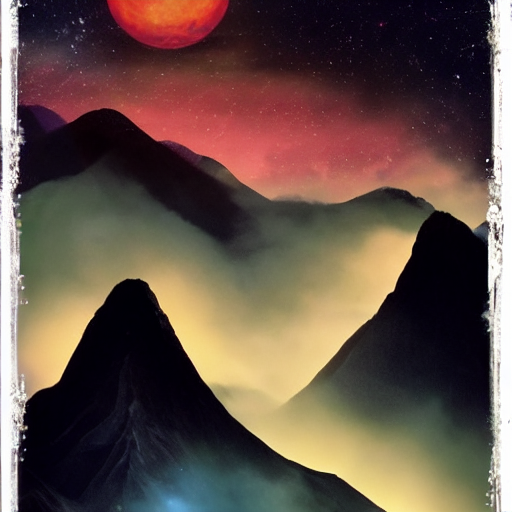}
        \caption{{\subB}}
    \end{subfigure}
    \hfill
    \begin{subfigure}[t]{0.43\textwidth}
        \centering
        \includegraphics[height=4.5cm]{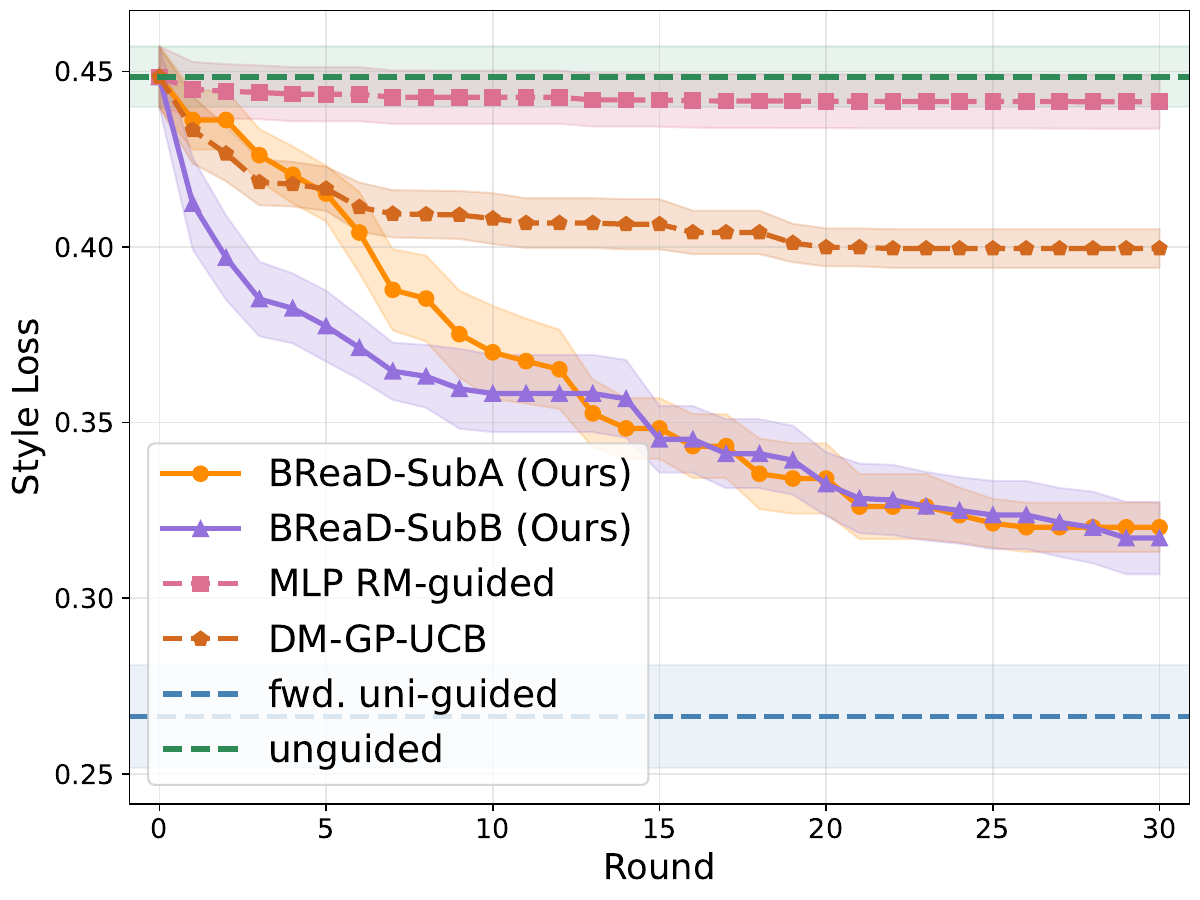}
        \caption{Best-so-far style loss}
    \end{subfigure}

    \caption{Style guidance on Stable Diffusion v1.5 to match the text prompt, \textit{A fantasy photo of an volcanoes}, and the style of Leonid Afremov's painting, \textit{Melody of The Night}. Results are averaged over 10 trials.}
    \label{fig:style guidance 2}
\end{figure}

\begin{figure}[!ht]
    \centering

    \begin{subfigure}[b]{0.25\textwidth}
        \centering
        \includegraphics[height=3.4cm]{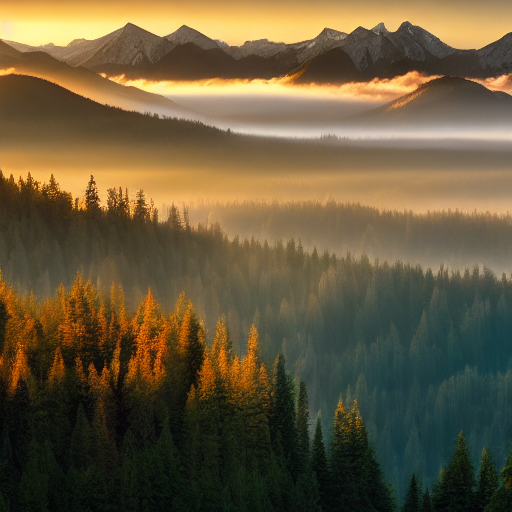}
        \caption{{\unguided} (Reward: \num{-0.94})}
        \label{fig:hpr guidance 3(a)}
    \end{subfigure}
    \hfill
    \begin{subfigure}[b]{0.25\textwidth}
        \centering
        \includegraphics[height=3.4cm]{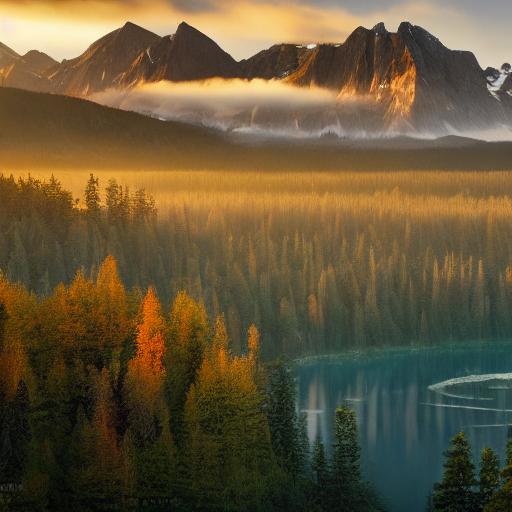}
        \caption{{\subB} (Round 11, Reward: \num{1.37})}
        \label{fig:hpr guidance 3(b)}
    \end{subfigure}
    \hfill
    \begin{subfigure}[b]{0.43\textwidth}
        \centering
        \includegraphics[height=4.5cm]{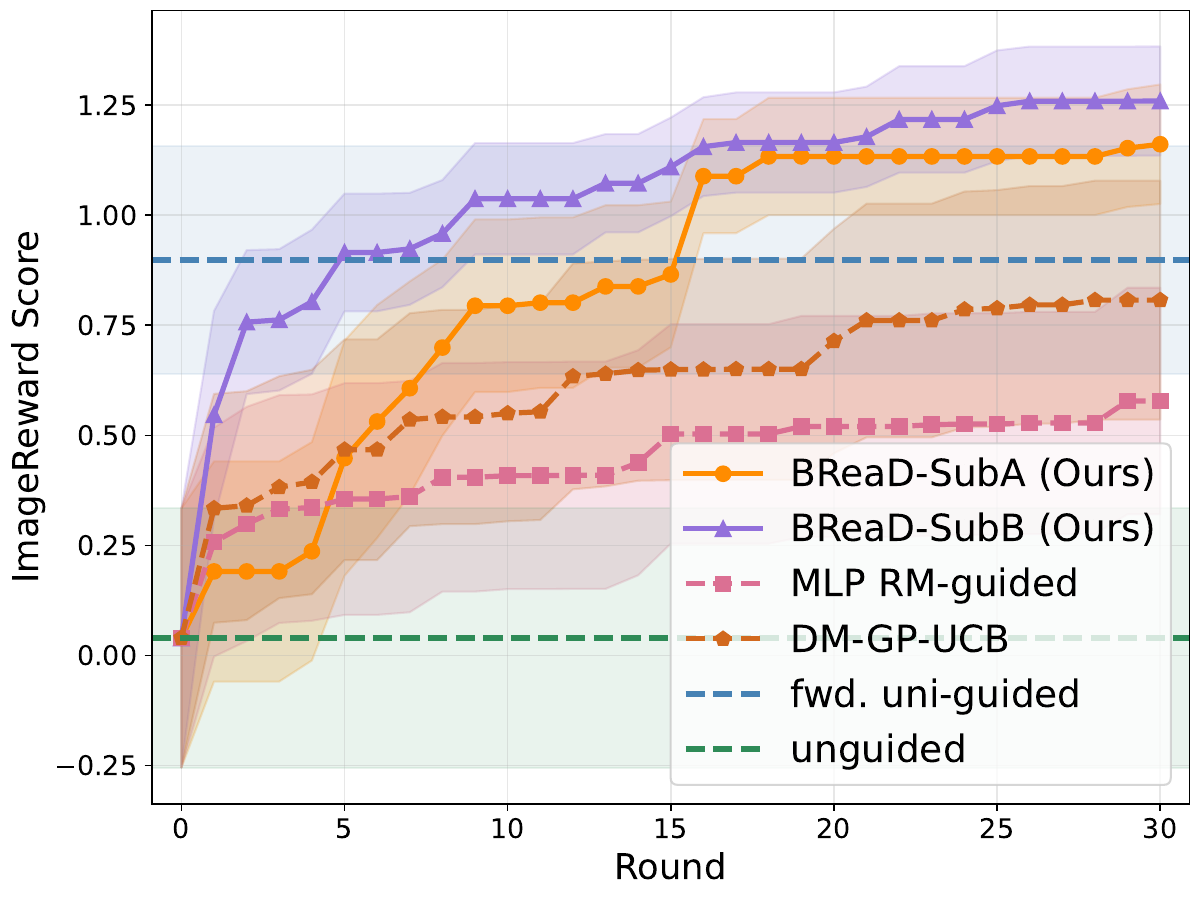}
        \caption{Best-so-far ImageReward score}
        \label{fig:hpr guidance 3(c)}
    \end{subfigure}

    \caption{Guidance sampling of Stable Diffusion v1.5 to match the text prompt, \textit{Landscape photography by Marc Adams, mountains with some forests, small lake in the center, fog in the background, sunrays, golden hour, high quality}, where the rewards are given by the ImageReward model. The images demonstrated use the same seed. Results are averaged over 10 trials.}
    \label{fig:hpr guidance 3}
\end{figure}

\begin{figure}[!h]
    \centering

    \begin{subfigure}[b]{0.31\textwidth}
        \centering
        \includegraphics[height=3.4cm]{figs/cold_hpr_sunflower.png}
        \caption{\textcolor{matplotlibforest}{unguided}}
    \end{subfigure}
    \quad
    \begin{subfigure}[b]{0.31\textwidth}
        \centering
        \includegraphics[height=3.4cm]{figs/round_011_fwd_sunflower.png}
        \caption{\textcolor{matplotsteelblue}{fwd. uni-guided}}
    \end{subfigure}
    \quad
    \begin{subfigure}[b]{0.31\textwidth}
        \centering
        \includegraphics[height=3.4cm]{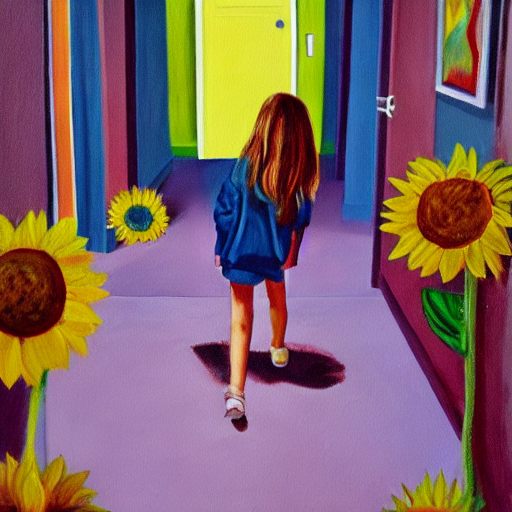}
        \caption{\textcolor{matplotpaleviolet}{MLP-guided}}
    \end{subfigure}
    \\
    \begin{subfigure}[b]{0.31\textwidth}
        \centering
        \includegraphics[height=3.4cm]{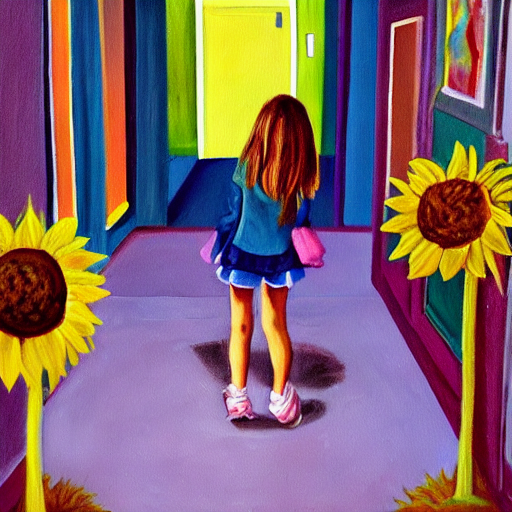}
        \caption{\gpucb}
    \end{subfigure}
    \begin{subfigure}[b]{0.31\textwidth}
        \centering
        \includegraphics[height=3.4cm]{figs/round_021_gp_sunflower.png}
        \caption{\textcolor{matplotdarkorange}{BReaD-SubA}}
    \end{subfigure}
    \quad
    \begin{subfigure}[b]{0.31\textwidth}
        \centering
        \includegraphics[height=3.4cm]{figs/round_011_ucb_hpr_sunflower.png}
        \caption{\textcolor{matplotmediumpurple}{BReaD-SubB}}
    \end{subfigure}
    \caption{Image generation for optimizing the ImageReward score. \textit{A painting of a girl walking in a hallway and suddenly finds a giant sunflower on the floor blocking her way.}}
    \label{fig:hpr guidance examples}
\end{figure}

\begin{figure}[!h]
    \centering
    \begin{subfigure}[t]{0.46\textwidth}
        \centering
        \includegraphics[height=4.8cm]{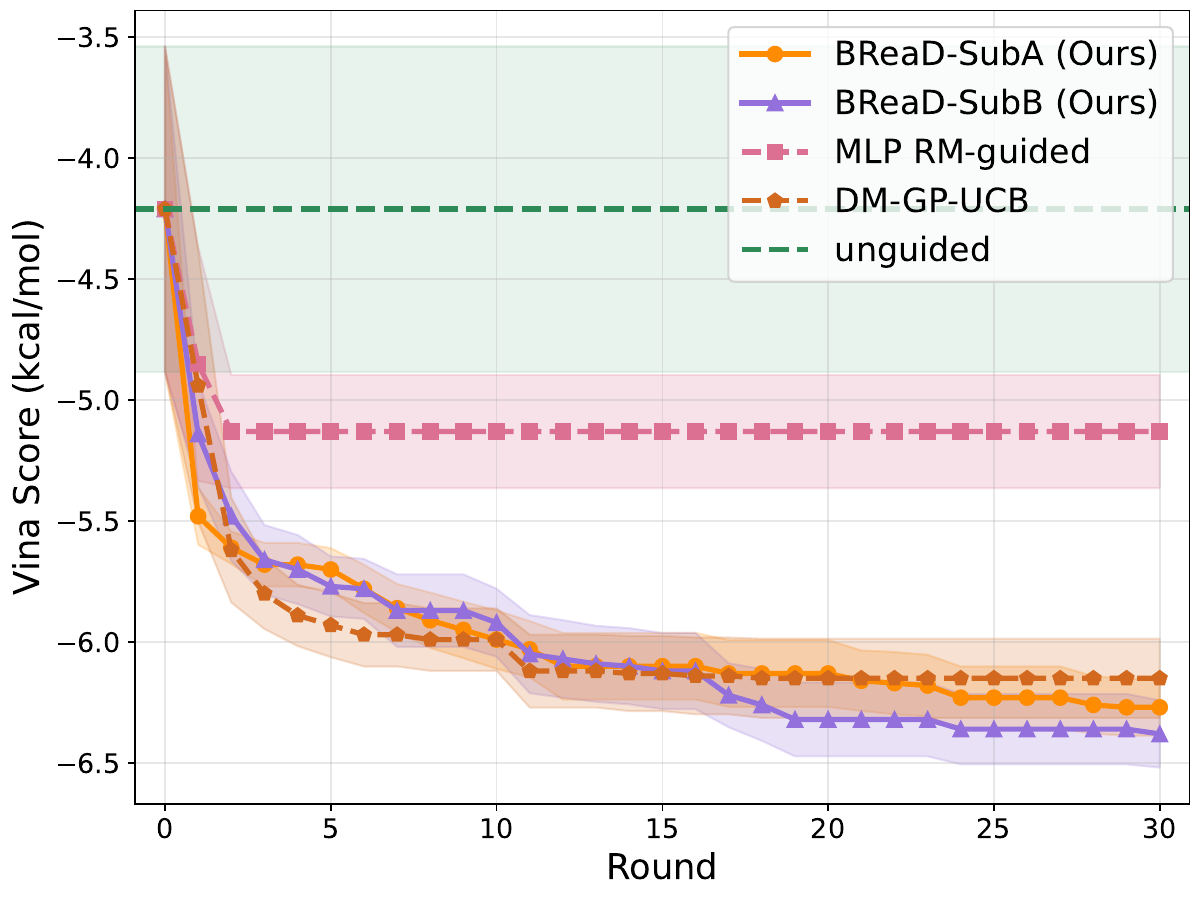}
        \caption{Vascular Endothelial Growth Factor Receptor 2 (VEGFR2)}
        \label{fig:prot_desgn_3}
    \end{subfigure}
    \quad
    \begin{subfigure}[t]{0.46\textwidth}
        \centering
        \includegraphics[height=4.8cm]{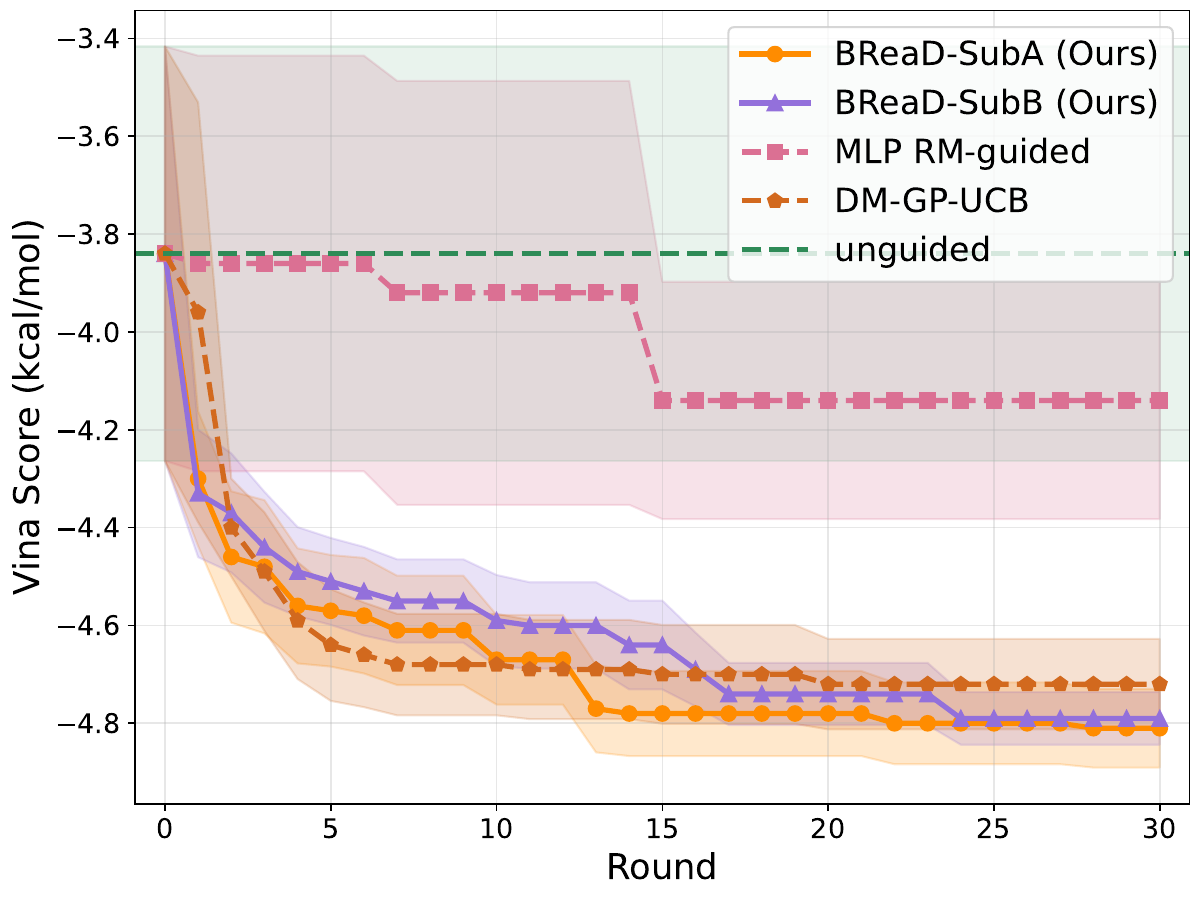}
        \caption{Estrogen receptor alpha (ER$\alpha$)}
        \label{fig:prot_desgn_4}
    \end{subfigure}
    \caption{\textbf{Ligand optimization.} VEGFR2 and ER$\alpha$. Results are averaged over \num{10} trials.}
    \label{fig:sci_discovery1}
\end{figure}

%% file: 2-related_work.tex
\textbf{High-dimensional Bayesian optimization.}\quad
Bayesian optimization (BO) is effective for optimizing expensive black-box objectives but often struggles in high-dimensional spaces. REMBO~\citep{wang2013rembo} assumes that the objective has low effective dimensionality and performs BO in a randomly embedded low-dimensional subspace. TuRBO~\citep{eriksson2019turbo} performs BO within local trust regions and adaptively expands or contracts these regions during optimization. SAASBO~\citep{eriksson2021saasbo} places a sparsity-inducing prior over input dimensions. BAxUS~\citep{papenmeier2022baxus} progressively scales up the dimensionality of the target subspace.

\textbf{Diffusion models and controlled generation.}\quad
A separate line of work uses pre-trained diffusion models to generate samples with desired properties. Classifier guidance~\citep{dhariwal2021diffusion} modifies the diffusion score using gradients from an auxiliary classifier, while classifier-free guidance~\citep{ho2021classifierfree} performs conditional generation by jointly learning the estimation of unconditional and conditional scores. Diffusion Posterior Sampling~\citep{chung2023diffusion} extends gradient-guided diffusion to inverse problems by incorporating gradients derived from an observation likelihood. Universal Guidance enables a pre-trained diffusion model to be controlled by arbitrary differentiable guidance functions without retraining task-specific components~\citep{bansal2024universal}. These approaches assume access to a differentiable function that provides guidance gradients during sampling, which are not readily available in black-box settings.

\textbf{Generative black-box optimization.} 
Generative models have increasingly been used to optimize black-box objectives in high-dimensional design spaces. \citet{pmlr-v97-brookes19a} propose adaptive sampling of generative models that estimates a distribution over the design
space conditioned on desired properties. Model inversion networks~\citep[MINs]{NEURIPS2020_373e4c5d} learn an inverse function that maps values to corresponding inputs. DDOM~\citep{pmlr-v202-krishnamoorthy23a} trains a reward-conditioned diffusion model from a large labeled dataset and generates candidates conditioned on a high objective value. Diff-BBO~\citep{wu2024diffbbo} further extends this approach to an online setting and uses an uncertainty-aware acquisition strategy to propose the conditioned objective. Reward-directed generation~\citep{yuan2023rewarddirected} uses a pseudo-labeler trained on labeled data to annotate an unlabeled dataset and then trains a reward-conditioned diffusion model. Recent work integrates Bayesian optimization into generative model-based optimization. GenBO~\citep{pmlr-v202-mashkaria23a} approximates a target distribution proportional to BO’s acquisition function to propose the next query locations. BONET~\citep{pmlr-v202-mashkaria23a} approximates the dynamics of black-box optimizers using an autoregressive sequence model. LaMBO~\citep{pmlr-v162-stanton22a} and LaMBO-2~\citep{NEURIPS2023_29591f35} jointly train a diffusion model with a discriminator and perform gradient-based optimization of an acquisition function. Fast Direct~\citep{tan2025fastdirect} updates a GP surrogate using batched black-box evaluations and constructs a pseudo-target for guidance. DAS~\citep{kim2025das} performs Sequential Monte Carlo~(SMC) during the reverse diffusion process to approximate a reward-tilted distribution at test time. DiBO~\citep{yun2025posterior} iteratively retrains and fine-tunes a diffusion model to sample from the posterior distribution induced by a deep-ensemble reward surrogate to balance exploration and exploitation. Multi-BO~\citep{rajagopalan2026personalized} performs BO on the diffusion model's attention features and iteratively uses human feedback to generate new candidates.